\documentclass[journal]{IEEEtran} 
\usepackage{amsmath,amsfonts}
\usepackage{algorithmic}
\usepackage{algorithm}
\usepackage{array}
\usepackage[caption=false,font=normalsize,labelfont=sf,textfont=sf]{subfig}
\usepackage{textcomp}
\usepackage{stfloats}
\usepackage{url}
\usepackage{verbatim}
\usepackage{cite}
\usepackage{booktabs}
\usepackage[margin=1in]{geometry}
\usepackage{graphicx}
\usepackage[colorlinks=true, linkcolor=blue, citecolor=blue, urlcolor=blue, hidelinks]{hyperref}
\DeclareUnicodeCharacter{039C}{M}
\usepackage[most]{tcolorbox}
\newtcolorbox{wipbox}{
  colback=red!10,
  colframe=red!75!black,
  boxrule=0.8mm,
  arc=1mm,
  left=2mm,
  right=2mm,
  top=1mm,
  bottom=1mm,
  box align=center,
  halign=center,
}

\begin{document}

\title{Raw2Event: Converting Raw Frame Camera into Event Camera}

\author{
    \IEEEauthorblockN{
        Z. Ning$^{\dag}$,
        E. Lin$^\S$,
        S.R Iyengar$^{\dag}$,
        P. Vandewalle$^{\dag}$
    }
    \IEEEauthorblockA{
        \\\textit{$^{\dag}$EAVISE-PSI, ESAT, KU Leuven, Belgium} \\
        \textit{$^\S$ESAT, KU Leuven, Belgium} \\
        zijie.ning@kuleuven.be
    }
}



\maketitle

\begin{abstract}
Event cameras offer unique advantages such as high temporal resolution, low latency, and high dynamic range, making them more and more popular for vision tasks under challenging light conditions. However, their high cost, limited resolution, and lack of features such as autofocus hinder their broad adoption, particularly for early-stage development and prototyping. In this work, we present Raw2Event, a complete hardware-software system that enables real-time event generation from low-cost raw frame-based cameras. By leveraging direct access to raw Bayer data and bypassing traditional image signal processors (ISP), our system is able to utilize the full potential of camera hardware, delivering higher dynamic range, higher resolution, and more faithful output than RGB-based frame-to-event converters.

Built upon the DVS-Voltmeter model, Raw2Event features a configurable simulation framework optimized for deployment on embedded platforms. We further design a data acquisition pipeline that supports synchronized recording of raw, RGB, and event streams, facilitating downstream evaluation and dataset creation. Experimental results show that Raw2Event can generate event streams closely resembling those from real event cameras, while benefiting from higher resolution and autofocus capabilities. The system also supports user-intuitive parameter tuning, enabling flexible adaptation to various application requirements. Finally, we deploy the system on a Raspberry Pi for real-time operation, providing a scalable and cost-effective solution for event-based vision research and early-stage system development.

The codes are available online\footnote{\href{https://anonymous.4open.science/r/raw2event-BFF2/README.md}{https://anonymous.4open.science/r/raw2event-BFF2/README.md}}.

\end{abstract}


\section{Introduction}
Frame-based computer vision has been extensively studied and adopted in various applications such as autonomous driving and robotics. However, traditional frame cameras suffer from fundamental limitations such as motion blur, and cannot perform well under challenging lighting due to their limited dynamic range. These issues introduce significant latency in object detection and decision-making. 
In safety-critical scenarios such as autonomous vehicles, such delays can be life-threatening. Therefore, it is essential to explore low-power, fast and adaptive vision systems that can meet the real-time demands of computer vision.

In recent years, event-based cameras, inspired by neuroscience, have emerged as a promising solution. These asynchronous sensors detect brightness changes at each pixel with microsecond latency, offering low power consumption, high dynamic range, and no motion blur. Their characteristics make them particularly well suited for embedded edge applications with limited computational and communication resources. As a result, event-based sensors have attracted increasing attention from both academia and industry, with active efforts dedicated to their integration into vision systems for applications such as robotics, surveillance, and autonomous driving~\cite{gallegoEventbasedVisionSurvey2022}.

However, commercial event cameras remain expensive, making them inaccessible for many developers. Additionally, these cameras often have low resolution and require manual calibration due to the absence of features like autofocus. Simulators such as v2e~\cite{huV2eVideoFrames2021}, v2ce~\cite{zhangV2CEVideoContinuous2024}, and DVS-Voltmeter~\cite{linDVSVoltmeterStochasticProcessBased2022} exist, but they typically run offline, lack real-time capability, and focus heavily on mimicking hardware-level behavior. This focus often neglects the practical need for user-controlled noise levels and tunable parameters, which are crucial in early-stage development and system prototyping. 

Furthermore, current frame-to-event simulation methods operate on RGB images or video sequences processed by ISP (Image Signal Processors), which inherently limit the accessible dynamic range and introduce artifacts such as compression noise and contrast distortion. Compression methods also result in different visual artifacts, including color shift, blocking, blurring, or ringing irregularities~\cite{jamilReviewImageQuality2024}.

To address these limitations, we propose Raw2Event, a real-time event generation system based on low-cost raw frame cameras. Leveraging the physics-driven event generation model introduced in DVS-Voltmeter~\cite{linDVSVoltmeterStochasticProcessBased2022}, Raw2Event allows developers to emulate the behavior of an event camera using affordable hardware such as a Raspberry Pi camera.

Our main contributions are as follows:

\begin{itemize}
  \item We introduce a configurable, real-time event generator based on raw pixel data. To our knowledge, we propose the first complete hardware-software event data simulation pipeline, including camera setup, raw data decoding, and real-time event conversion.
  \item We use the raw frame stream before processing by traditional ISPs, supporting higher dynamic range, and saving more information from the scene.
  \item We extend the work of DVS-Voltmeter~\cite{linDVSVoltmeterStochasticProcessBased2022} by proposing a more accurate parameter calibration method using search, and give human-understandable meanings to the parameters. We also provide a guide on tuning these parameters and understanding their impact.
  \item To our knowledge, we propose the first pipeline that simultaneously collects both raw frame data and event data at the same time. This includes sensor-robot communication, customized motion control, cross-sensor time synchronization and automated post-processing, facilitating dataset collection and future research on event data generator.

\end{itemize}

The rest of this paper is organized as follows: Section~\ref{ch:RelatedWork} reviews related work in event-based vision, simulation tools, and datasets. Section~\ref{ch:SystemDesign} presents the overall system design, including the motivation, hardware setup, software pipeline, and synchronized data acquisition. Section~\ref{ch:eventgeneration} details the configurable event generation model and the parameter calibration strategy. Section~\ref{ch:resultsandevaluation} evaluates the proposed system through qualitative comparisons, parameter tuning studies, and real-time performance. Finally, Section~\ref{ch:conclusionandfuturework} concludes with a discussion of future work and potential extensions of Raw2Event.

\section{Related Work} 
\label{ch:RelatedWork}

\subsection{Event Based Vision}

Traditional frame-based cameras operate by capturing full-intensity images at fixed frame rates, irrespective of the actual dynamics occurring within the scene. While this approach has been highly successful across a wide range of computer vision applications, it faces several fundamental limitations when applied to high-speed or high-dynamic-range environments. First, frame-based sensors are susceptible to motion blur when objects move rapidly within the exposure time window~\cite{Deblurring}. Second, they possess a limited dynamic range, making it challenging to capture fine details of the scene under extreme lighting conditions~\cite{hdr}. Third, the continuous acquisition of full image frames results in the processing of large amounts of redundant information, unnecessarily consuming bandwidth and computational resources~\cite{lichtsteiner128times1281202008}.

Event-based cameras, such as the Dynamic Vision Sensor (DVS)~\cite{lichtsteiner128times1281202008}, fundamentally depart from this conventional sensing paradigm. Instead of recording full frames at fixed intervals, event cameras asynchronously output streams of events, where each event is independently triggered by a pixel in response to a significant change in local brightness. Each event encodes the pixel coordinates, the polarity of the brightness change, and a precise timestamp. Owing to this sensing mechanism, event cameras achieve exceptionally high temporal resolution and ultra-low latency, making them particularly suited for capturing fast-moving scenes. Their inherently high dynamic range enables robust performance in challenging illumination conditions. Furthermore, the sparse and efficient nature of event data allows computational resources to focus solely on meaningful scene changes rather than redundant static information~\cite{Gehrig2024Nature}, significantly reducing the burden on data transmission and processing systems.

Driven by these advantages, an increasing number of companies and research groups have started to explore the use of event cameras to develop intelligent vision applications~\cite{mentastiEventBasedObjectDetection2022,falangaDynamicObstacleAvoidance2020,jiangMixedFrameEventDriven2019}. However, the high cost of commercial event cameras remains a significant barrier, especially for startups and early-stage projects where budget constraints limit experimentation. This situation underscores the need for a low-cost, accessible solution to facilitate the development and testing of event-based vision applications without the immediate investment in specialized hardware.

\subsection{Event Camera Simulator}

Event camera simulators have become essential tools for the development and evaluation of event-driven visual algorithms. Despite their unique advantages, DVS have not yet achieved widespread adoption. Unlike frame-based vision, which benefits from a wealth of available datasets such as ImageNet~\cite{imagenet}, no large-scale, standardized event datasets have been created so far. Therefore, researchers often rely on simulation tools to generate synthetic event data or convert frame-based datasets into event datasets.

The primary goal of event simulators is to approximate the behavior of real DVS pixels by converting frame sequences or rendered scenes into asynchronous event streams. These simulators can be broadly categorized into two types: threshold-based simulators, which generate events by comparing pixel intensity changes across frames against a threshold; and learning-based simulators, which treat event generation as a data-driven mapping problem and solve it using optimization methods or adversarial approaches.

Several prominent threshold-based simulators have been proposed, each introducing novel improvements in timestamp generation, event triggering logic, noise modeling, and computational performance:

\begin{itemize}
    \item DAVIS Simulator~\cite{muegglerEventCameraDatasetSimulator2017} was one of the earliest efforts to provide both an event camera dataset and a simulator. It follows a rule-based model that relies on logarithmic luminance differences and linear interpolation between video frames to generate events.
    \item ESIM~\cite{rebecqESIMOpenEvent2018} builds upon this concept by introducing an efficient pixel-level interpolation strategy. Instead of performing global upsampling, ESIM only interpolates pixel values that are likely to generate events, thus improving computational efficiency.
    \item v2e (Video to Events)~\cite{huV2eVideoFrames2021} simulates the behavior of real DVS pixels by introducing non-ideal characteristics such as pixel-level threshold mismatches, leakage events, and intensity-dependent bandwidth limitations.
    \item DVS-Voltmeter~\cite{linDVSVoltmeterStochasticProcessBased2022} models the event generation process as a random drift-diffusion process based on the internal voltage dynamics of DVS pixel circuits. Events are triggered when the modeled signal crosses a predefined contrast threshold. This leads to a hitting-time distribution that captures the temporal randomness and light-dependency observed in real sensors. Compared to previous simulators based on linear interpolation and fixed thresholds (e.g., v2e~\cite{huV2eVideoFrames2021}, Vid2E~\cite{gehrigVideoEventsRecycling2020}), DVS-Voltmeter offers a more realistic model in terms of temporal distribution and noise generation.
\end{itemize}

As for learning-based simulators, they have gained popularity as they can model complex, nonlinear mappings from frames to event streams. Researchers have explored methods in adversarial learning, supervised learning, and unsupervised learning for event simulation:

\begin{itemize}
    \item EventGAN~\cite{zhuEventGANLeveragingLarge2019} formulates the image-to-event conversion as an adversarial task. The model generates event streams from image pairs using a GAN framework.
    \item Nehvi et al.~\cite{nehviEventHandsRealTimeNeural2021} propose a differentiable rendering-based simulator for non-rigid 3D tracking. The simulator synthesizes events through a fully differentiable pipeline, enabling gradient-based optimization via backpropagation. This aligns with the convex optimization and gradient descent approaches widely used in deep learning.
    \item DA4Event~\cite{planamenteDA4EventBridgingSimtoReal2021} addresses the sim-to-real domain gap by applying unsupervised domain adaptation techniques to simulate event data (e.g., from ESIM~\cite{rebecqESIMOpenEvent2018}) and improve generalization on real-world event streams.
    \item V2CE~\cite{zhangV2CEVideoContinuous2024} introduces a deep learning framework that generates continuous event streams from video input, avoiding discrete frame assumptions and aiming to model the continuous temporal characteristics of DVS signals.
    \item Gu et al.~\cite{guReliableEventGeneration2024} present a learning-based reversible event generation model. Its core lies in using Conditional Normalizing Flow (CNF) to establish a bidirectional mapping between event streams and generation parameters (such as contrast threshold and noise rate).
\end{itemize}

Both methods have their advantages and disadvantages. Threshold-based simulators provide higher interpretability and fine control over simulation parameters, which is crucial for research and benchmarking. However, they may suffer from limited generalization across different camera models or illumination conditions. In contrast, learning-based simulators offer powerful data-fitting capabilities and can model complex correlations in real data. Yet, they often face challenges such as poor interpretability and high training costs.

In this work, we focus on real-time deployment and configurability, which naturally favors threshold-based methods. Raw2Event aims to strike a balance between realism, flexibility, and efficiency. Unlike learning-based simulators, which typically depend on pre-recorded videos and intensive training, Raw2Event operates directly on real-time raw sensor data, enabling event generation without relying on large-scale datasets or offline training.

We also note that existing event simulation frameworks typically focus on the conversion from RGB frames to events, either in the software or hardware domain. However, few studies provide a complete pipeline integrating camera setup, raw data acquisition, and real-time event conversion on accessible hardware. In this paper, we introduce the first complete simulation pipeline that links low-cost hardware with efficient event generation software. Our system processes raw sensor data directly, enabling more accurate brightness measurement and retaining a greater bit-depth range.

\subsection{Event Classification and Detection Datasets}
\label{sec:Event Classification and Detection Datasets}

With the rapid advancement of event-based vision technologies, numerous datasets designed specifically for tasks such as object detection and image classification have emerged. These datasets vary significantly in terms of resolution, recording methodology, and application domain. However, capturing new event-based datasets from scratch typically requires specialized DVS hardware, considerable resources, and substantial time investment. This makes custom event-data collection often impractical for researchers. Consequently, many widely used event datasets have been generated by converting conventional frame-based image datasets using real event cameras.

Several notable event-based datasets derived from their frame-based counterparts have been widely adopted. For instance, MNIST-DVS~\cite{serrano-gotarredonaPokerDVSMNISTDVSTheir2015} and N-MNIST~\cite{orchardConvertingStaticImage2015} were among the first event datasets, created by displaying MNIST handwritten digits on a screen. In these cases, they flash digits on the screen or move an event camera to create brightness changes. CIFAR10-DVS~\cite{liCIFAR10DVSEventStreamDataset2017} similarly converted the CIFAR-10~\cite{cifar10} dataset by recording event data as images moved across a display monitored by a DVS. A more comprehensive dataset, N-ImageNet~\cite{kimNImageNetRobustFineGrained2021}, adopted a similar recording strategy using the extensive ImageNet~\cite{imagenet} dataset. They kept the image stationary on the screen while the event camera rotated.

However, these varying acquisition methods often introduce unintended artifacts associated with hardware constraints. For example, the CIFAR10-DVS dataset contains bursts of events synchronized with the monitor's refresh rate, and the diagonal image translation strategy results in events being significantly concentrated at the vertices of the motion path. To mitigate such recording artifacts, researchers often choose to shuffle event timestamps. 
While such temporal shuffling may be acceptable for rate-based methods or temporal accumulation models, it inevitably discards potentially valuable temporal information, which can be particularly critical for spike-based neural networks and other temporally-sensitive algorithms. Therefore, it is crucial that event-based datasets closely reflect realistic temporal dynamics. 

Furthermore, most existing event datasets either contain only event data (e.g., DVS Gesture, N-Caltech101) or originate from pre-recorded frame-based inputs without direct access to raw sensor information. This limitation prevents researchers from modifying event generation parameters to simulate diverse event camera models or varied lighting conditions. Additionally, the use of ISPs and image compression in frame-based data collection results in quality degradation and loss of information.

Real-world event datasets captured directly from event cameras do exist~\cite{perotLearningDetectObjects2020,plizzariE2GOMOTIONMotionAugmented2022}. However, such datasets are currently limited in number and tend to focus on specific application areas like autonomous driving or egocentric action recognition. Although these datasets offer highly realistic event data directly captured from real-world scenarios, their limited availability and specialized nature pose difficulties for researchers working outside these specific fields. Thus, researchers often face challenges in acquiring real-world event data suited to their unique requirements.

Given these challenges, converting widely used frame-based datasets into event-based representations remains a practical and efficient alternative. The proposed Raw2Event system directly addresses these limitations, offering researchers a flexible platform to generate realistic event streams from raw frame-based images. This approach eliminates the need for specialized event-camera hardware, enabling researchers to quickly construct, evaluate, and train event-based models across diverse application scenarios. Additionally, our platform allows researchers to adjust simulation parameters easily, facilitating the emulation of various hardware characteristics and environmental conditions. To further support comprehensive dataset creation, we have also developed an integrated data acquisition pipeline that simultaneously collects raw sensor data, RGB images, and event streams, allowing further research and experimentation.

\section{System Design}
\label{ch:SystemDesign}

\subsection{Motivation and Design Principles}

The primary goal of Raw2Event is not to replicate the extreme sensitivity or microsecond-level temporal resolution of event cameras, as these capabilities are inherently constrained by the physical limitations of traditional frame-based image sensors. Instead, our goal is to offer a practical and accessible alternative for typical application scenarios, where ultra-high frame rates or extreme sensitivity in darkness are not required. In many practical applications, such as autonomous driving, where motion is within the speed range manageable by conventional frame-based cameras and lighting conditions are favorable (daytime or illuminated roads), their frame rate and dynamic range are typically sufficient.

In addition to the cost-related advantage, frame-based cameras provide several practical benefits: smaller physical size, significantly higher resolution (e.g., 4K compared to 346×260 for DAVIS346), and the availability of features like autofocus, which are rarely supported in commercial event sensors.

Moreover, the benefits of converting frame-based inputs into event-like data have been widely discussed in the literature~\cite{gallegoEventbasedVisionSurvey2022,chakravarthiRecentEventCamera2024}. Major advantages include reduced bandwidth consumption, as only meaningful changes are transmitted, and enhanced privacy protection, as not the entire image is transmitted. Such representations allow downstream tasks to focus on the changes in the scene, avoiding the redundant processing of unchanging information.

In this sense, Raw2Event is a smart combination of frame and event cameras, combining the accessibility and flexibility of frame cameras with the temporal efficiency of event-based representations. It also allows the development of an event processing pipeline without the immediate need for expensive event camera hardware.

Furthermore, the use of raw sensor data introduces additional advantages. It allows us to bypass the ISP, giving direct access to each pixel's unprocessed intensity values. This avoids interference from automatic exposure adjustments, noise reduction, color correction, and other factors which can distort the temporal brightness changes that are critical for event generation. Also, raw data typically provides higher bit depth under the same hardware design, which translates to higher dynamic range in output.

\subsection{Hardware Setup}

Modern visual sensing systems offer a variety of camera technologies beyond conventional RGB sensors. In addition to standard frame cameras, commonly used hardware includes high-speed cameras, low-light night vision sensors, and thermal or infrared imagers. Each type serves different application needs and exhibits trade-offs in terms of cost, resolution, dynamic range, and performance. While alternative modalities such as LiDAR and radar are also used in perception systems, this work focuses solely on vision-based approaches.

To support cost-efficient event simulation, our system employs the Raspberry Pi Camera Module 3 as the image acquisition device. This compact and affordable module integrates the Sony IMX708 image sensor and supports up to 12MP resolution, High Dynamic Range (HDR) mode, and phase detection autofocus (PDAF).

\begin{figure}[htbp]
    \centering
    \includegraphics[width=0.9\columnwidth]{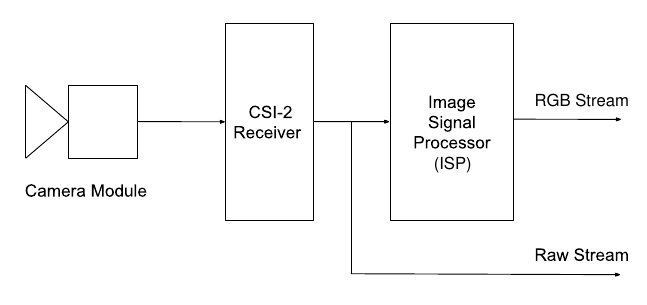}
    \caption{Hardware connection diagram.~\cite{picamera2manual} The camera module transmits image data via the CSI-2 receiver to the ISP, which outputs both the processed RGB stream and the raw image stream.}
    \label{fig:camera_stream}
\end{figure}

A key feature of the Raspberry Pi platform is that it provides official support for raw sensor access via drivers and tools such as “picamera2”, which allow us to capture unprocessed Bayer-format data before it reaches the ISP. This direct access to raw data allows Raw2Event to bypass ISP stages such as demosaicing, tone mapping, denoising, and automatic gain control, all of which are typically designed to enhance visual appearance for human viewers, but may distort per-pixel intensity changes. Many off-the-shelf cameras for embedded systems only provide ISP-processed output, limiting their utility in low-level vision modeling. In contrast, the Pi camera is capable of providing both the raw stream and the main RGB stream simultaneously, enabling comparative analysis and controlled experiments on the influence of the ISP. The hardware connection diagram of the Raspberry Pi Camera Module is shown as in Fig.~\ref{fig:camera_stream}, illustrating how image signals split into raw and ISP-processed streams for event generation and conventional vision tasks respectively. 

Beyond raw pixel values, the Camera Module 3 also exposes rich sensor metadata, including exposure time, gain, temperature, and timestamps, which are instrumental for event triggering, noise modeling, and temporal alignment in our simulation pipeline.

The Raspberry Pi camera is connected to an embedded compute module (in our case, Raspberry Pi), which communicates with a host PC via a local Ethernet network, ensuring high-throughput and low-latency data transfer suitable for real-time or batch processing scenarios.

\subsection{Software Pipeline}

\begin{figure}[htbp]
    \centering
    \includegraphics[width=0.9\linewidth]{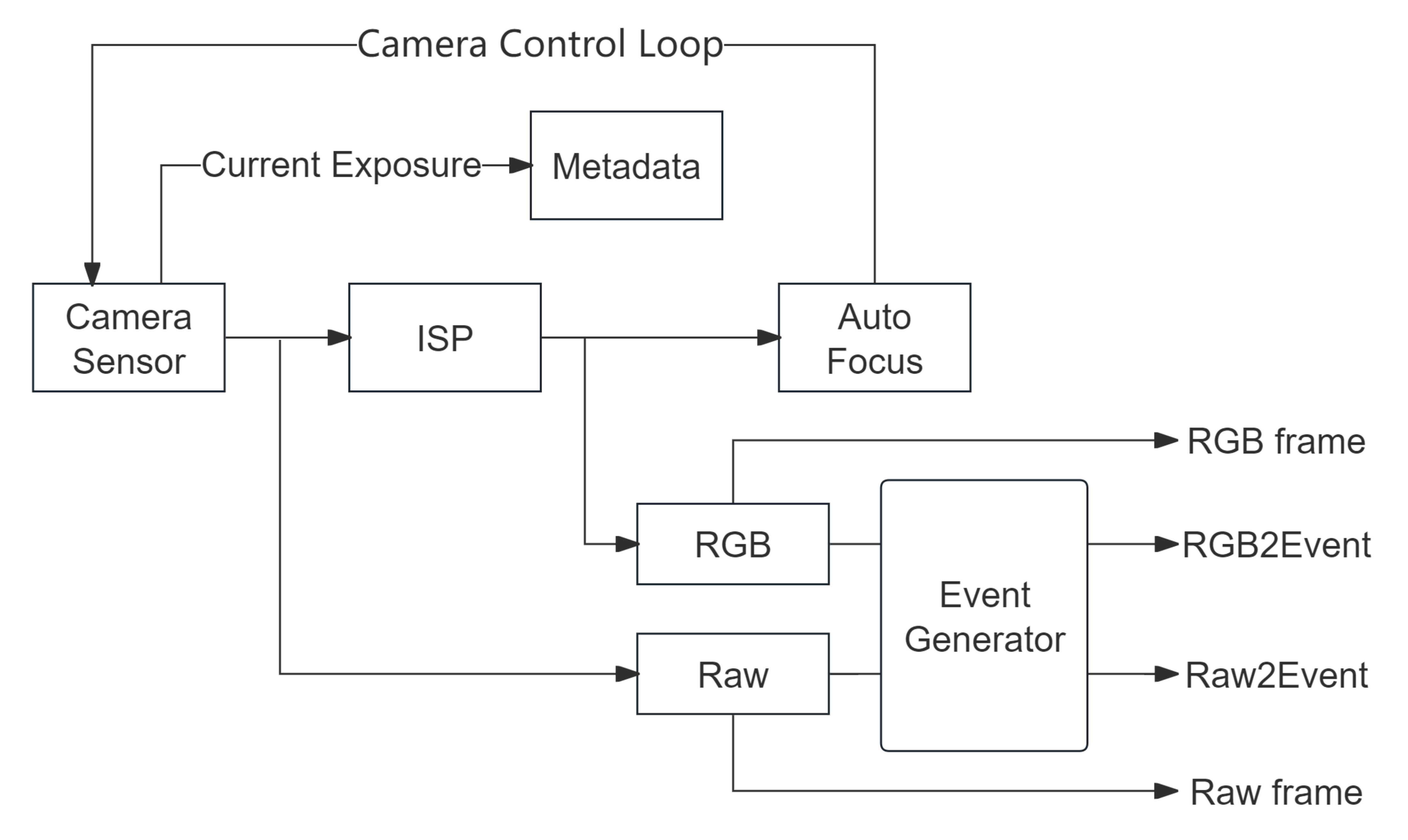}
    \caption{Software data flow and control loop of the Raw2Event system. The camera sensor simultaneously outputs both raw Bayer data and ISP-processed RGB frames. These are fed into the \textit{Event Generator} module, which produces two synthetic event streams: Raw2Event and RGB2Event. Both raw and RGB frames can also be stored for offline analysis. The camera’s internal control loop manages parameters via metadata feedback, while we only enable autofocus to ensure consistent pixel intensity dynamics.}
    \label{fig:software_flow}
\end{figure}

Fig.~\ref{fig:software_flow} illustrates the overall data flow and camera control loop of the Raw2Event system. The software pipeline is designed not only to support real-time event stream generation, but also to store raw and RGB frame data for offline analysis and reproducibility. Specifically, both the raw Bayer frames and ISP-processed RGB frames are simultaneously saved, and independently passed into the Event Generator module described in Section~\ref{ch:eventgeneration}. This yields two parallel outputs: Raw2Event, generated from 10-bit raw intensity values; and RGB2Event, based on the corresponding 8-bit RGB stream.

While Raw2Event typically yields higher fidelity and is the preferred mode for downstream applications, the preservation of both RGB frames and RGB2Event data streams provides a basis for calibration, ablation studies, and comparative research on the effects of ISP transformations.

Camera control is managed via an internal feedback loop that dynamically adjusts parameters such as exposure time, analog/digital gain, and autofocus, based on scene statistics. In our implementation, the automatic exposure and gain control algorithms are disabled, and the exposure time is fixed to avoid introducing variability into pixel-level brightness values. Only the autofocus function remains active in the loop, as it does not directly interfere with raw pixel intensities. At the same time, it offers a significant practical advantage over current event cameras, which typically lack autofocus function. This contributes to sharper image content, which can be beneficial for both visualization and downstream tasks.


The system can be run offline, capturing raw video data and later converting it into event streams, utilizing the parallel computing capabilities of GPUs. When deploying Raw2Event in time-sensitive applications, such as robotics or embedded vision systems, real-time performance is an essential consideration. In this context, the achievable frame rate becomes a critical performance metric that directly impacts the system's responsiveness. The original DVS-Voltmeter simulator relies on PyTorch-based GPU acceleration, which is not feasible on embedded platforms such as the Raspberry Pi. To accommodate this constraint, we re-implement the core event generation logic using NumPy-based vectorized operations. This design enables the system to run on lightweight hardware with acceptable frame rates for real-time prototyping and evaluation, even in the absence of specialized hardware accelerators.

\subsection{Synchronized Acquisition System for Frame and Event Data}
\label{sec:synchronized_acquisition}

As mentioned in Section~\ref{sec:Event Classification and Detection Datasets}, current event datasets often offer only event streams or frames from the internal grayscale sensors of event cameras, without supplementary high-resolution RGB images, raw frame data, or accurately recorded motion trajectories. To enable accurate evaluation of Raw2Event and address this gap, we built a desktop-scale acquisition system capable of spatially and temporally synchronized capture of both frame-based and event-based data under the same precisely controlled motion trajectory.

\subsubsection{System Setup}

\begin{figure}[htbp]
    \centering
    \includegraphics[width=0.9\columnwidth, trim=0 0 0 0, clip]
    {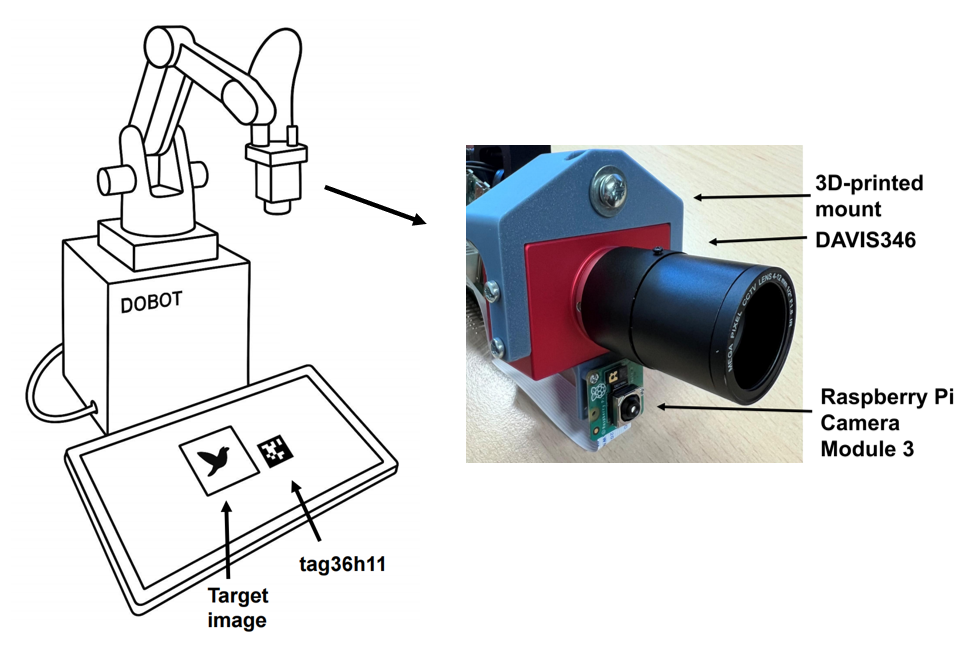}
    \caption{Setup of the Dobot, DAVIS346, Pi Camera Module 3, target image, the AprilTag and the 3D-printed dual-camera mount.}
    \label{fig:system}
\end{figure}

Our acquisition platform integrates a Dobot Magician 6-DoF robotic arm and a custom 3D-printed dual-camera mount as shown in Fig.~\ref{fig:system}. The mount holds a DAVIS346 mono event camera and a Raspberry Pi Camera Module 3 at the same horizontal height, positioning the two lenses side by side. This setup ensures both cameras observe the scene from similar perspectives.

\subsubsection{Cross-Sensor Time Synchronization}
Accurate cross-sensor timestamp alignment is essential to mitigate the influence of variable factors across different experimental sessions. The DAVIS346 camera can simultaneously record real event data and grayscale frames, both accompanied by Unix-based timestamps that can be directly mapped to real-world time, thus ensuring natural internal synchronization. In contrast, the Raspberry Pi records timestamps relative to its own system clock, which bears no inherent relation to real-world time, making cross-sensor synchronization with the DAVIS346 challenging. To address this, we record two types of timestamps for each frame in the Raspberry Pi metadata: the SensorTimestamp, extracted from the camera’s internal metadata and representing the precise hardware exposure time; and the RealTime, captured via \texttt{datetime.now()}, which represents the current global time synchronized via Network Time Protocol (NTP) through internet. 

\begin{figure}[htbp]
    \centering
    \includegraphics[width=0.95\columnwidth]{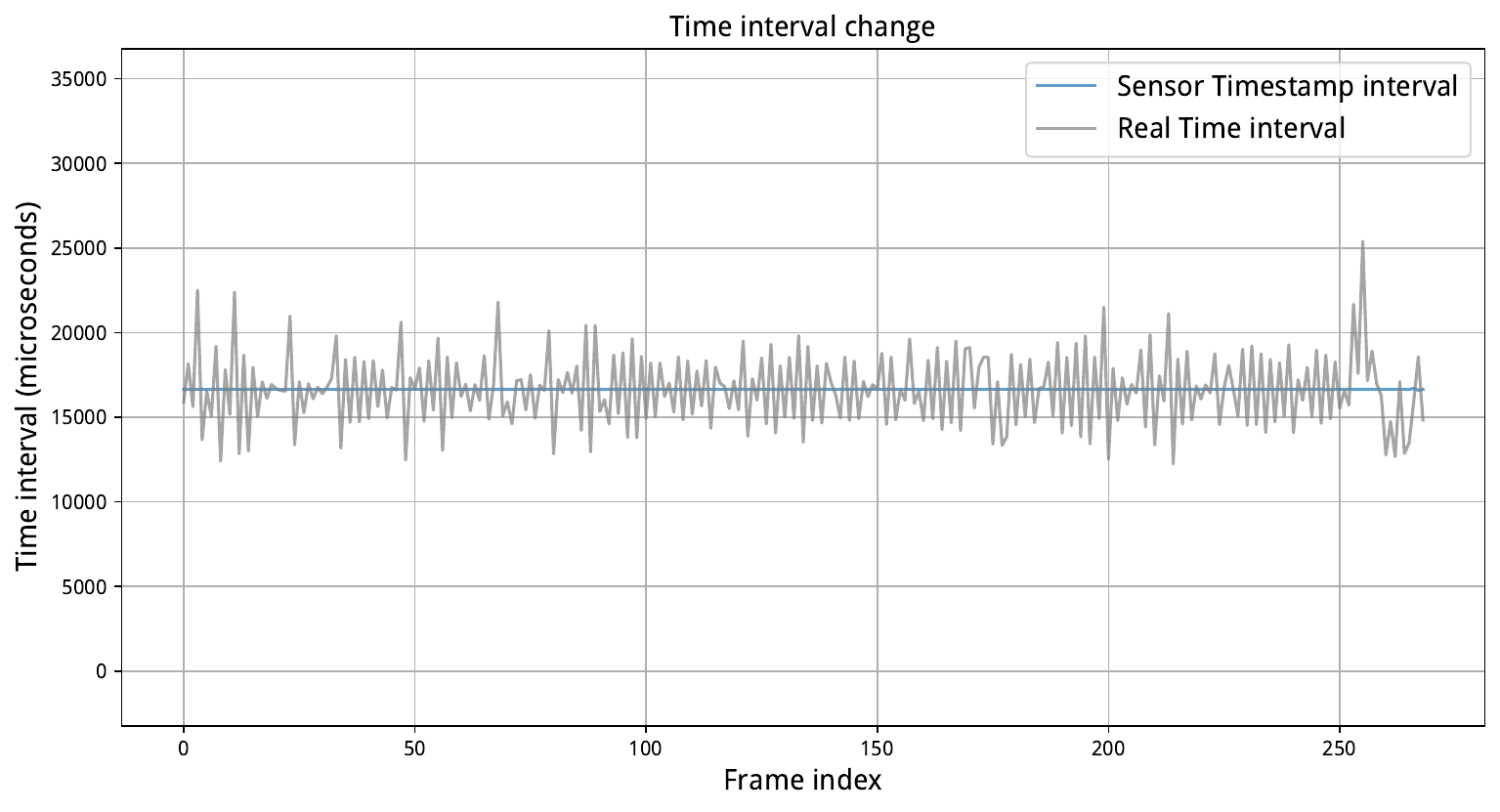}
    \caption{Time interval changes over frames.}
    \label{fig:time_intervals}
\end{figure}

Precise synchronization between devices depends on estimating the time offset between the SensorTimestamp and RealTime clocks. As shown in Fig.~\ref{fig:time_intervals}, the SensorTimestamp offers high temporal accuracy and consistent frame intervals. In contrast, the RealTime values show significant jitter due to API call latency. However, over a sufficiently long observation window, the average frame interval of the RealTime sequence tends to converge toward the stable interval seen in the SensorTimestamp sequence. We identify the moment where the two interval sequences are approximately equal and use the difference between their timestamp values at this moment to estimate the offset between SensorTimestamp and RealTime. With this offset, we align the Raspberry Pi's internal timestamps to the real-world timestamps used by the DAVIS346.

\subsubsection{Controlled Circular Motion}

Unlike many existing datasets that employ polygonal trajectories composed of straight segments (e.g., triangular, rectangular, or diamond-shaped paths), we adopt a smooth circular trajectory. Sharp turns inherent in polygonal paths often cause concentrated bursts of events due to unstable directional transitions. Such bursts result in uneven temporal distributions. Moreover, directional occlusion can occur in linear motion, where edges aligned with the movement direction produce insufficient brightness changes to trigger events. In contrast, a circular path enables continuous angular change and avoids such periodic artifacts, resulting in temporally smoother and statistically more consistent event data. Furthermore, as mentioned in Section~\ref{sec:Event Classification and Detection Datasets}, moving the camera over static displays is a more realistic and reliable strategy to generate event data, as it can avoid artifacts introduced by monitor refresh mechanisms and preserve more realistic time information.

\subsubsection{ROI Tracking and Event Filtering}

\begin{figure*}[htbp]
  \centering
  \subfloat[DVS Camera Position Trajectory\label{fig:dv_trajectory}]{%
    \includegraphics[width=0.32\textwidth]{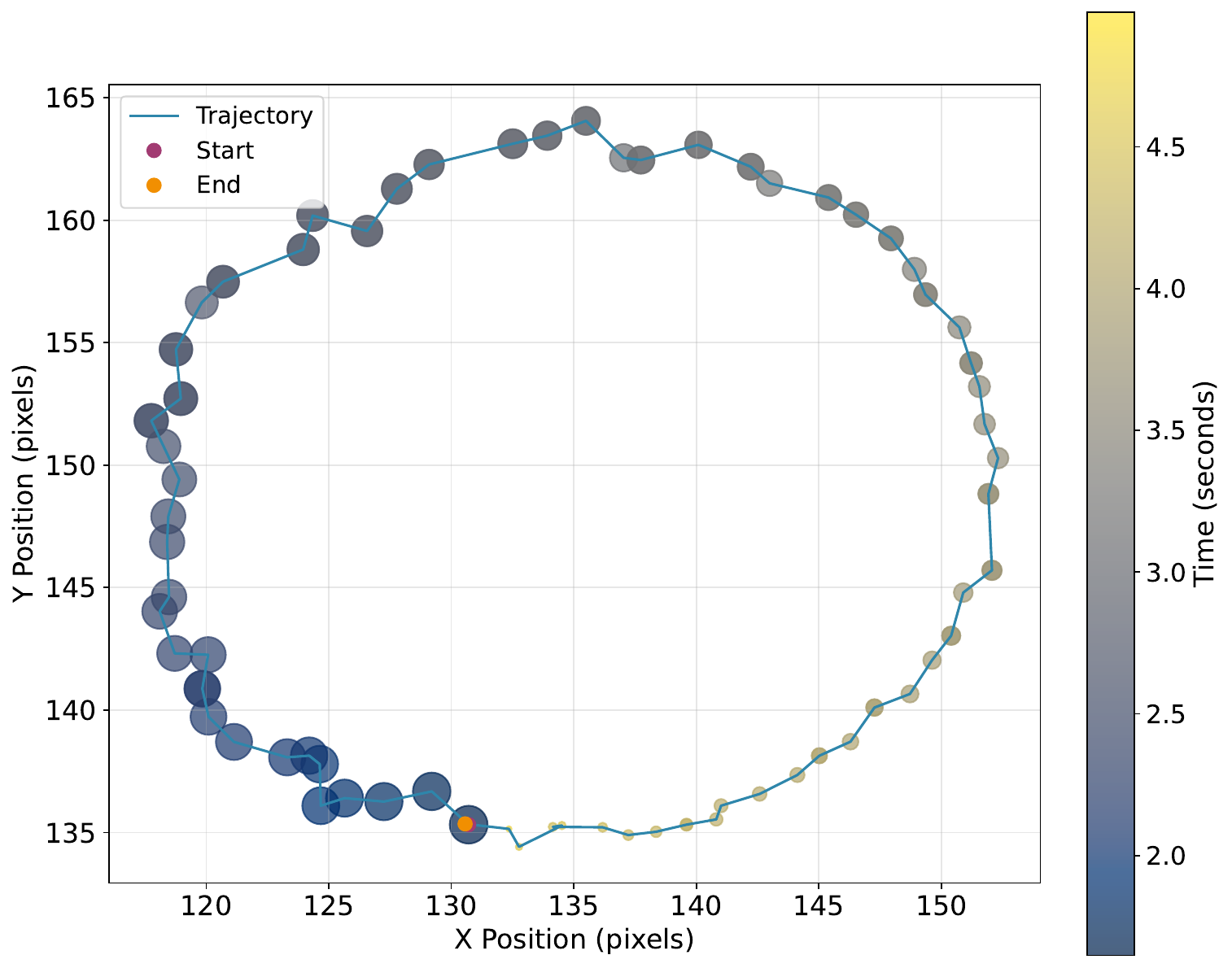}%
  }\hfill
  \subfloat[Rotation Angle Comparison\label{fig:angle_comparison}]{%
    \includegraphics[width=0.32\textwidth]{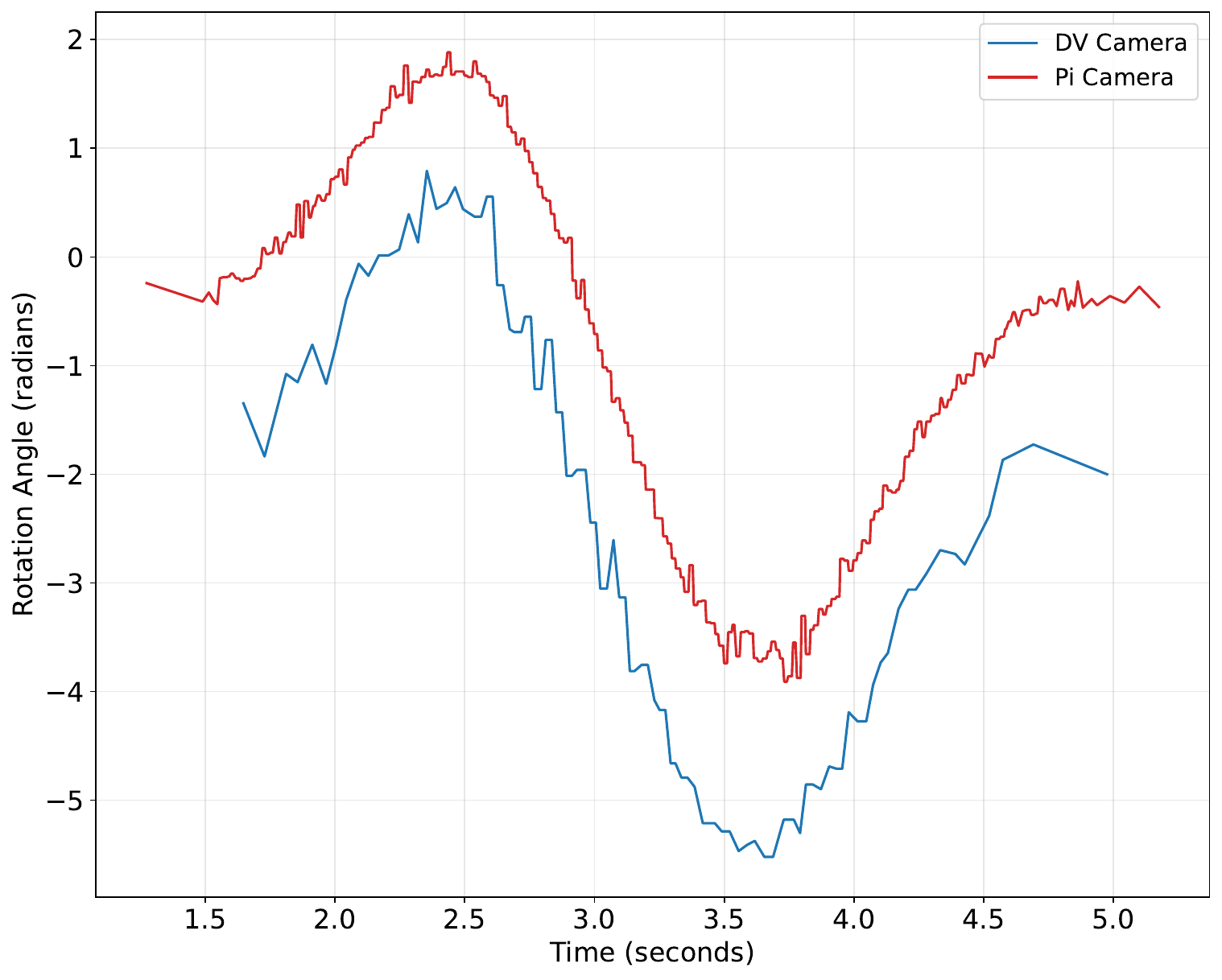}%
  }\hfill
  \subfloat[Pi Camera Position Trajectory\label{fig:pi_trajectory}]{%
    \includegraphics[width=0.32\textwidth]{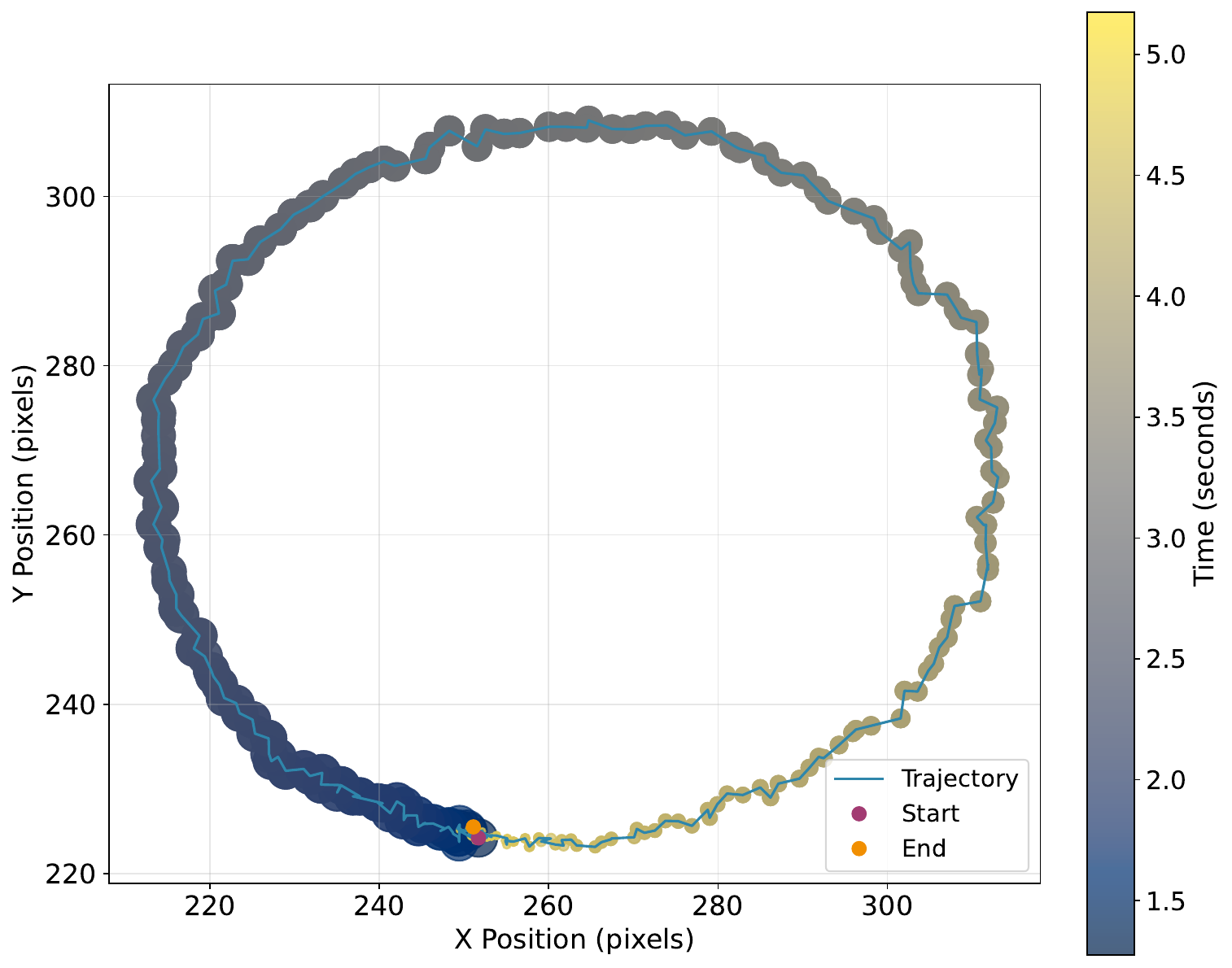}%
  }%
  \caption{Comparison of rotation angle and spatial trajectories between DVS and Pi camera data.}
  \label{fig:all_comparison}
\end{figure*}

To track the object of interest while the camera is moving, we place an AprilTag~\cite{wangAprilTag2Efficient2016} adjacent to the target image. The known relative size and position between the tag and the target allows us to track the size, position, and rotation angle of the target region in each frame. This method ensures robust tracking of the region of interest (ROI) throughout the motion sequence, providing resilience against mechanical jitter and other perturbations, and making it suitable for any desired trajectories. We extract only the events that occur within the tracked ROI in both real and simulated data streams. For both real DAVIS346 events and simulated Raw2Event events, the corresponding grayscale frames are used to detect the tag and determine the ROI. This unified tracking and filtering pipeline ensures consistent spatial alignment and supports direct comparison between the two sources. An example of the trajectory of the tracked ROI center and its rotation angle over time during circular camera motion can be seen in Fig.~\ref{fig:all_comparison}.

\section{Configurable Event Generation}
\label{ch:eventgeneration}

Based on the analysis in Section \ref{ch:RelatedWork}, we choose to use the event generation model introduced by DVS-Voltmeter~\cite{linDVSVoltmeterStochasticProcessBased2022}, which provides one of the most realistic physics-aware models of DVS behavior. While our work does not introduce new event simulation theory, our contribution lies in utilizing the physically meaningful parameters within this model to provide users with explicit, fine-grained control over the simulated event generation process. By systematically investigating these parameters, we enable the configuration of both event and noise phenomena inherent to event-based cameras. This transforms the simulation from a fixed process into a versatile tool, bridging the gap between raw frame input and tailored, realistic event data output configurable by the user. 

This section is organized as follows: Section~\ref{sec:Event Generation Model and Noise Characterization} reviews the underlying event generation model and emphasize the role of its parameters. Section~\ref{sec:Cross-Sensor Parameter Calibration} presents the cross-sensor calibration methodology, allowing generalization to diverse input frame camera and output event sensor. Finally, Section~\ref{sec:User-Intuitive Parameters for Event Behavior Control} describes how the calibrated parameters are exposed as user-tunable controls.

\subsection{Event Generation Model and Noise Characterization}
\label{sec:Event Generation Model and Noise Characterization}

The DVS-Voltmeter approach models the change in a pixel's internal logarithmic photoreceptor voltage, $\Delta V_d$, over a short time interval $\Delta t$. Its strength lies in formulating this change by incorporating multiple physical effects, including signal transduction, photon shot noise, and various leakage currents. The comprehensive model explicitly considers the noise contributions from leakage current (influenced by temperature and parasitic photocurrent), as well as signal transduction and shot noise, relates $\Delta V_d$ to the scene brightness ($\bar{L}$), brightness change rate ($k_{dL}$), and sensor-specific parameters such as temperature that are captured by $k_1$ to $k_6$. The model can be expressed as a stochastic differential equation, where $W(\Delta t)$ denotes a Wiener process increment: 

\begin{align}
\label{eq:delta_vd_withk}
\Delta V_d = & \left( \frac{k_1}{\bar{L} + k_2} k_{dL} + k_4 + k_5 \bar{L} \right) \Delta t \nonumber \\
&+ \left( \frac{k_3}{\bar{L} + k_2} \sqrt{\bar{L}} + k_6 \right) W(\Delta t).
\end{align}

We can clearly see that this event generation model can effectively be summarized as a Brownian motion term with drift $\mu$ and diffusion $\sigma$:

\begin{equation}
\label{eq:delta_vd}
    \Delta V_d = \mu \Delta t + \sigma W(\Delta t)
\end{equation}

Events (ON/OFF) are generated when $\Delta V_d(t)$ first reaches the respective thresholds ($-\Theta_{ON}, +\Theta_{OFF}$), with the event timestamp $\tau$ sampled probabilistically from the corresponding first hitting time which follows an Inverse Gaussian distribution.

This physics-based structure allows the model to inherently capture and distinguish several key phenomena contributing to the drift ($\mu$) and scale ($\sigma$) terms, crucial for realistic event simulation:
\begin{itemize}
    \item \textit{Photon Shot Noise} (via $k_3$): The inherent quantum randomness in photon arrivals contributes to $\sigma$. This effect is proportional to $\sqrt{L}$ (and modulated by $k_2$), introducing more temporal jitter in brighter regions.
    \item \textit{Temperature-Dependent Leakage Drift} (via $k_4$): A constant drift term. It simulates baseline leakage currents that are primarily influenced by temperature. It results in background events (often of ON polarity) independent of light intensity, and is noticeable in static dark areas.
    \item \textit{Parasitic Photocurrent Leakage Drift} (via $k_5$): An additional drift component proportional to brightness ($L$) is represented by $k_5$. This models leakage induced by incident light, increasing the background event rate specifically in brighter static regions.
    \item \textit{Baseline Electronic/Leakage Noise} (via $k_6$): This represents the sensor's noise floor (e.g., thermal, quantization noise), adding a baseline level of noise independent of light or motion.
\end{itemize}

In essence, the model links the characteristics of the event stream to these underlying physical effects quantified by the tunable parameters $k_1$ through $k_6$. This decomposition provides the foundation for user control over the simulation output.

\subsection{Cross-Sensor Parameter Calibration}
\label{sec:Cross-Sensor Parameter Calibration}

As shown in Eq.~(\ref{eq:delta_vd_withk}), the voltage change \(\Delta V_d\) follows a Brownian motion with drift. According to classical stochastic process theory, the first hitting time \(\tau\)—the time when the process hits a threshold—follows an inverse Gaussian distribution, shown as below:

\begin{equation}
\label{eq:IG distribution}
\tau_{\text{ON}}, \tau_{\text{OFF}} \sim 
\begin{cases}
IG\left( \frac{\mp\Theta_{\text{ON/OFF}}}{\mu}, \frac{\Theta_{\text{ON/OFF}}^2}{\sigma^2} \right), & \mu \neq 0 \\
\text{Levy}\left( \mu, \frac{\Theta_{\text{ON/OFF}}^2}{\sigma^2} \right), & \mu = 0
\end{cases}
\end{equation}
This corresponds to the event data as follows: the time interval \(\tau = t_{i+1} - t_{i}\), where \(E_i(t_i, x_{i}, y_{i}, p)\) and \(E_{i+1}(t_{i+1}, x_{i}, y_{i}, p)\), is defined as the difference in timestamps between two successive events at the same pixel.

The drift $\mu$ and diffusion $\sigma$ are determined by the scene brightness and brightness change rate, defined as:

\begin{equation}
\label{eq:u_equation}
\mu = \frac{k_1}{L + k_2} k_{dL} + k_4 + k_5 L
\end{equation}

\begin{equation}
\sigma = \frac{k_3}{L + k_2} \sqrt{L} + k_6
\end{equation}
Therefore, \(\mu\) and \(\sigma\) parameters derived from {\(k_1, k_2, \dots, k_6\)} ultimately determine the distribution of \(\tau\), which serves as a crucial metric to assess whether the simulated event data matches the real timestamp behavior.

\subsubsection{Motivation for Sensor-Specific Calibration}
\label{sec:Motivation for Sensor-Specific Calibration}

The authors of DVS-Voltmeter~\cite{linDVSVoltmeterStochasticProcessBased2022} proposed a calibration method to fit the parameters \(\left\{ k_1, k_2, \dots, k_6 \right\}\).
These $k_i$ are sensor-related parameters and need to be calibrated for each pair of frame sensor and targeted event sensor.

\subsubsection{Data Preparation}

Specifically, we leverage our desktop-scale synchronized acquisition system described in Section~\ref{sec:synchronized_acquisition}, which captures temporally and spatially aligned data across all sensors 

The temporally aligned data enables us to locate its adjacent frames $F_{i}$ and $F_{i+1}$ for each event. The spatial alignment further allows us to extract corresponding luminance values $L_{i}$ and $L_{i+1}$ at the event's $(x,y)$ coordinates. For RGB data, we convert pixel values to grayscale using \texttt{cv2.COLOR\_BGR2GRAY}, while for raw data we directly utilize the native 10-bit pixel values.

We extend the standard event representation $(t, x, y, p)$ with derived luminance features, creating an enriched 6-tuple format:
\begin{equation}
    (t, x, y, p, \bar{L}, \Delta L)
\end{equation}
where $\bar{L} = (L_{i} + L_{i+1})/2$ is the approximated brightness and $\Delta L = L_{i+1} - L_{i}$ is the brightness change.

\subsubsection{Grouping and Histogram Fitting}
For each brightness pair $(\bar{L}, \Delta L)$, we then perform a statistical histogram analysis of the event time intervals. Inverse Gaussian fitting is applied to estimate the drift $\mu$ and diffusion $\sigma$. If the brightness mean is close to zero, L\'evy distribution fitting is used to describe the event timing distribution without drift. We then obtain a set of data points in the form of \(\{\mu, \sigma, \bar{L}, \Delta L, N\}\), where \(N\) denotes the number of samples in each brightness pair, for use in the three-step regression (Section.~\ref{sec:Three-step Regression}).

To obtain sufficient $N$ for histogram statistics, we group brightness pairs $(\bar{L}, \Delta L)$ into two-dimensional intervals by uniformly dividing the ranges of mean brightness and brightness difference. Each interval aggregates events falling within its bounds and is represented by the midpoint of its range. Intervals with insufficient samples are discarded.

\subsubsection{Three-step Regression}
\label{sec:Three-step Regression}
For each data point, the brightness change rate is computed as \(k_{dL} = \Delta L / \Delta t\), with \(\Delta t = 1 / f\) denoting the frame interval determined by frame rate \(f\). At this point, we have all the data for $k$-parameter linear regression. The fitting of Eq.~(\ref{eq:u_equation}) is divided into three regression stages:
\begin{itemize}
\item First Regression (Grouped Linear Regression): A weighted linear regression is performed within each \(\bar{L}\) to fit the relationship between \(\mu\) and \(k_{dL}\): 
\begin{equation}
\mu = a_n \cdot k_{dL} + b_n
\end{equation}

where the weight assigned to each sample is the number of events in that interval \(N\), and \(a_n\), \(b_n\) denote the fitted slope and intercept, respectively. We updated to \(\{\mu, \sigma, \bar{L}, \Delta L, N,a_n, b_n\}\).

\item Second Regression (Global Linear Regression): There is a linear relationship between $1/a_n$ and $\bar{L}$:
\begin{equation}
\frac{1}{a_n} = \frac{1}{k_1} \bar{L} + \frac{k_2}{k_1}
\end{equation}
By performing weighted regression over all data, we estimate a single set of global parameters \(\{k_1,k_2\}\).
\item Third Regression (Multivariate Linear Regression): Define $c_n = k_{dL}/(\bar{L} + k_2)$, then perform multivariate linear regression:
\begin{equation}
\mu = k_1' \cdot c_n + k_5 \cdot \bar{L} + k_4
\end{equation}
again using weighted regression. Parameters $k_4$, $k_5$, and $k_1'$ are obtained, where $k_1'$ should be close to $k_1$, providing a consistency check.
\end{itemize}

These regressions provide a useful estimation of the parameters $k$.

\subsubsection{Black-box Optimization}

To further refine the simulation quality, we incorporate a black-box optimization strategy using Optuna~\cite{optuna_2019}, a framework for efficient hyperparameter search.

We formulate the parameter search as a minimization problem, where the objective is to reduce the Earth Mover’s Distance (EMD) between the simulated event point cloud and the corresponding ground truth events captured by the DAVIS346 sensor. The EMD metric reflects the spatial and temporal distribution similarity between two event streams, providing a meaningful measure of fidelity.

\begin{figure}[htbp]
    \centering
    \includegraphics[width=0.9\linewidth]{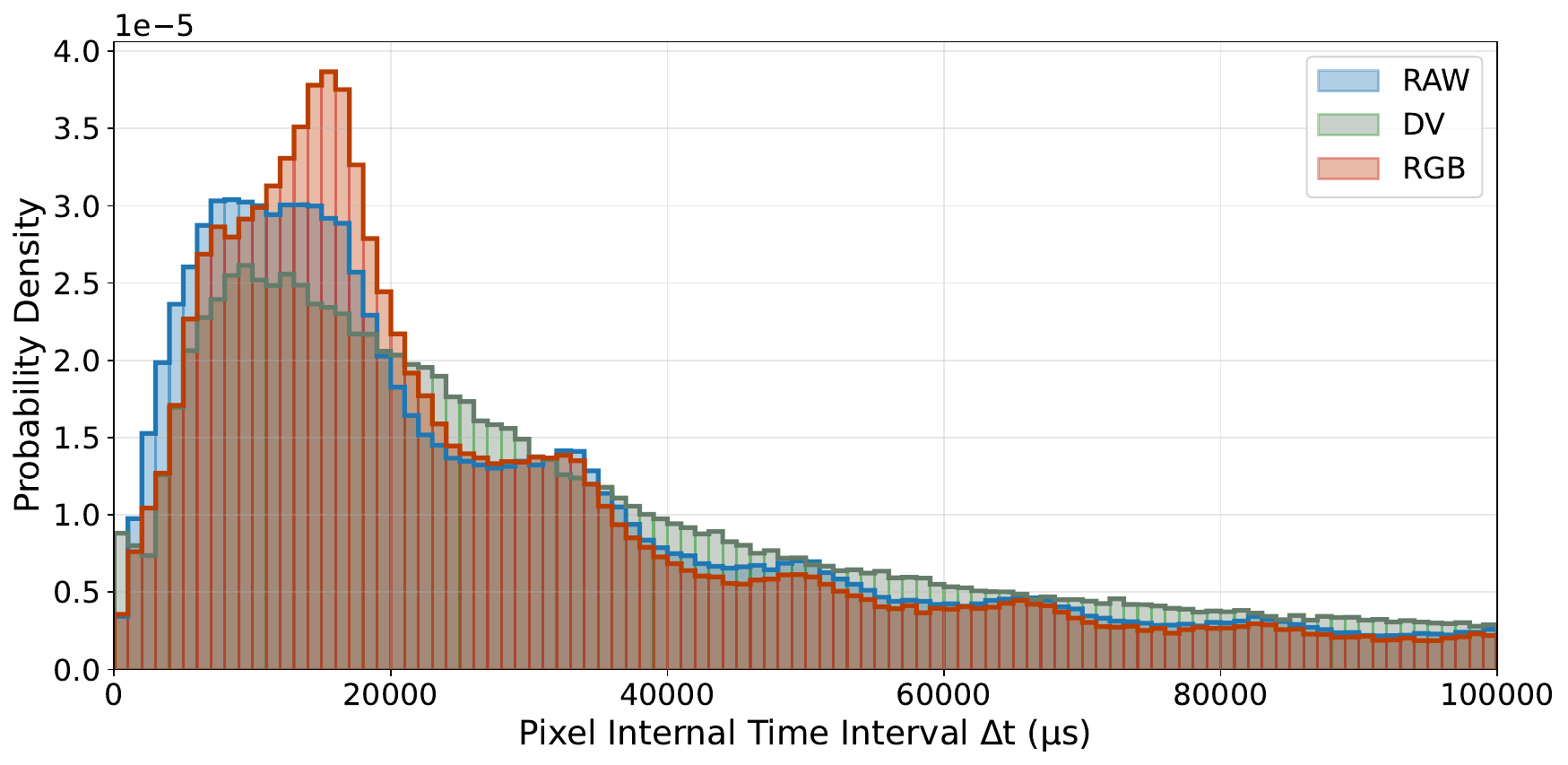}
    \caption{Probability density distribution of the internal time intervals between events for different data sources. The “RAW” and "RGB" lines represent the event stream generated from raw or RGB frame inputs using parameters optimized via Optuna. The “DV” line corresponds to the ground-truth events recorded by the DAVIS346 sensor.}
    \label{fig:k_calib_heatmap}
\end{figure}

Fig.~\ref{fig:k_calib_heatmap} illustrates the probability density distribution of the internal time intervals between events for different data sources. As seen in the figure, the distribution of both Raw2Event (raw-data-based simulation) and RGB2Event (RGB-data-based simulation) approximate the pattern of DAVIS346, demonstrating both temporal smoothness and accurate event density. The RGB2Event diverges more from the event camera distribution than Raw2Event, especially in high-frequency regions, due to its limited dynamic range and ISP-induced distortions. These results validate the effectiveness of above-mentioned $k$ calibration pipeline. We will use these $k$ values as baseline for following experiments.

\subsection{User-Intuitive Parameters for Event Behavior Control}
\label{sec:User-Intuitive Parameters for Event Behavior Control}

Building on the physical interpretations of parameters $k_1 \dots k_6$, our contribution is to expose these physically motivated quantities as direct user-facing controls. While the raw parameters $k$ capture underlying physics, they lack immediate intuition to configure specific simulation behaviors. To address this, we provide a mapping between the $k$ values and observable characteristics of the generated event stream, enabling users to configure event generation according to their specific requirements. This can be used for example to emulate a particular event camera or to optimize the event characteristics for downstream algorithms.

Table \ref{tab:event_params} provides an overview of these relationships, linking user-centric descriptions of behavior to the underlying $k$ parameters and highlighting the expected effects of parameter tuning. 

\begin{table*}[htbp]
\centering
\small
\renewcommand{\arraystretch}{1.3}
\setlength{\tabcolsep}{5pt}
\begin{tabular}{|>{\raggedright\arraybackslash}p{2.7cm}
                |>{\raggedright\arraybackslash}p{2.7cm}
                |>{\raggedright\arraybackslash}p{2.9cm}
                |>{\raggedright\arraybackslash}p{3.2cm}
                |>{\raggedright\arraybackslash}p{2.7cm}|}
\hline
\textbf{User-facing Parameter} & \textbf{Control Purpose} & \textbf{Definition (in terms of \(k\))} & \textbf{Effect Description} & \textbf{Recommended Parameter Range} \\
\midrule
Event Sensitivity/ Exposure & Response to brightness variations & \(k_1\) (modulated by \(k_2\)) & Higher → easier to trigger events; Lower → less responsive & $k_1$:1 to 10 \\
Contrast Sensitivity & Ability to detect edges & \(k_1\) and \(k_4\) & Higher → detects subtle edges easily; Lower → higher activation threshold for event generation & $k_1$: 1 to 10 ; $k_4$: 0 to \(10^{-7}\)\\
Bright Leakage & Events in static bright areas & \(k_5\) & Higher → static bright areas will fire events more frequently & $k_5$: 0 to \(10^{-7}\) \\
Temporal Jitter & Overall temporal randomness in event firing & \(k_3\) (light-dependent jitter) and \(k_6\) (baseline jitter) & Lower → decrease the number of spurious events; Higher → increases randomness in event timestamps & Not Applicable \\
Dark Noise Floor & Noise in dark or still background & \(k_4\) & Higher → more background noise in dark or still regions & $k_4$: 0 to \(10^{-7}\) \\
Global Noise Level & Overall random event noise & \(k_6\) & Higher → random noise across entire scene & $k_6$: 0 to \(10^{-4}\) \\
\bottomrule
\end{tabular}
\vspace{2pt}
\caption{User-intuitive parameter definitions for simulated event cameras based on calibrated model parameters \(k_1\) to \(k_6\), with their respective ranges (for the case in our research).}
\label{tab:event_params}
\end{table*}

\section{Results and Evaluation}
\label{ch:resultsandevaluation}

\subsection{Qualitative Comparison}

\subsubsection{High Dynamic Range Scene}

\begin{figure}[htbp]
    \centering
    \includegraphics[width=0.9\linewidth, trim=0 0 0 400, clip]{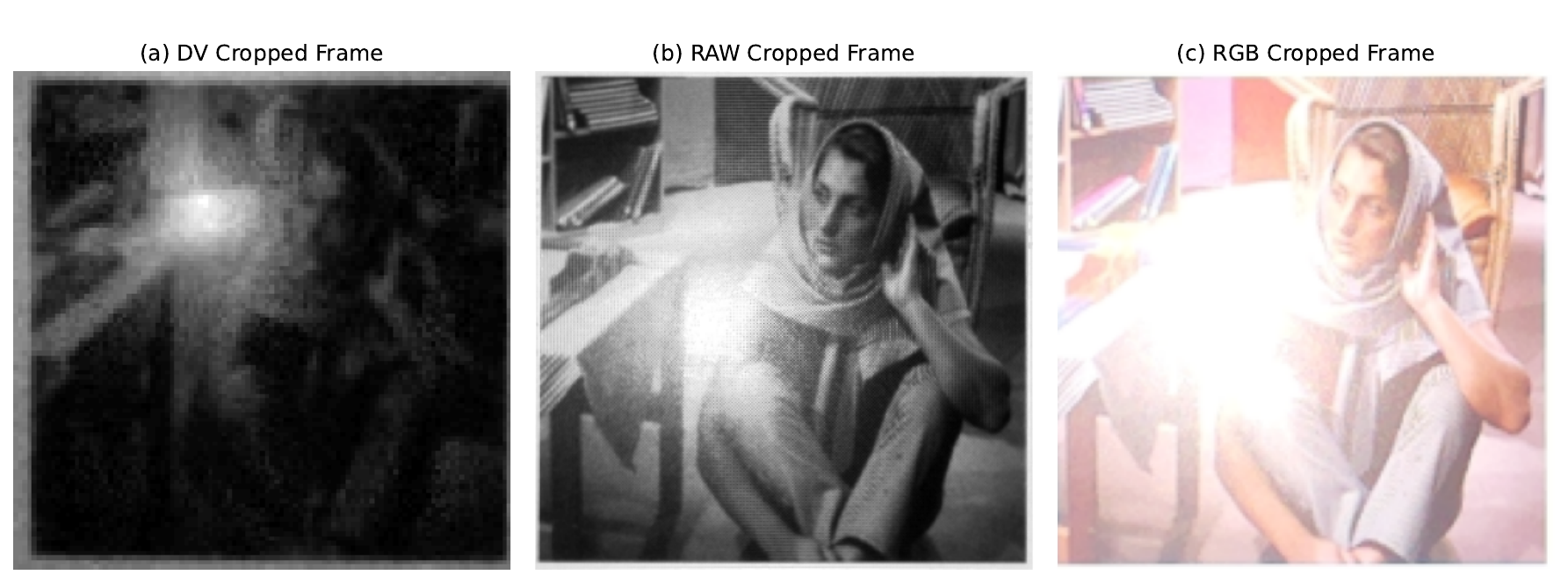}
    \caption{Views from frame sensor of DAVIS346, raw and RGB sensor of Pi Cam, under HDR scene.}
    \label{fig:hdr_crop_frame}
\end{figure}

\begin{figure}[htbp]
    \centering
    \includegraphics[width=0.9\linewidth]{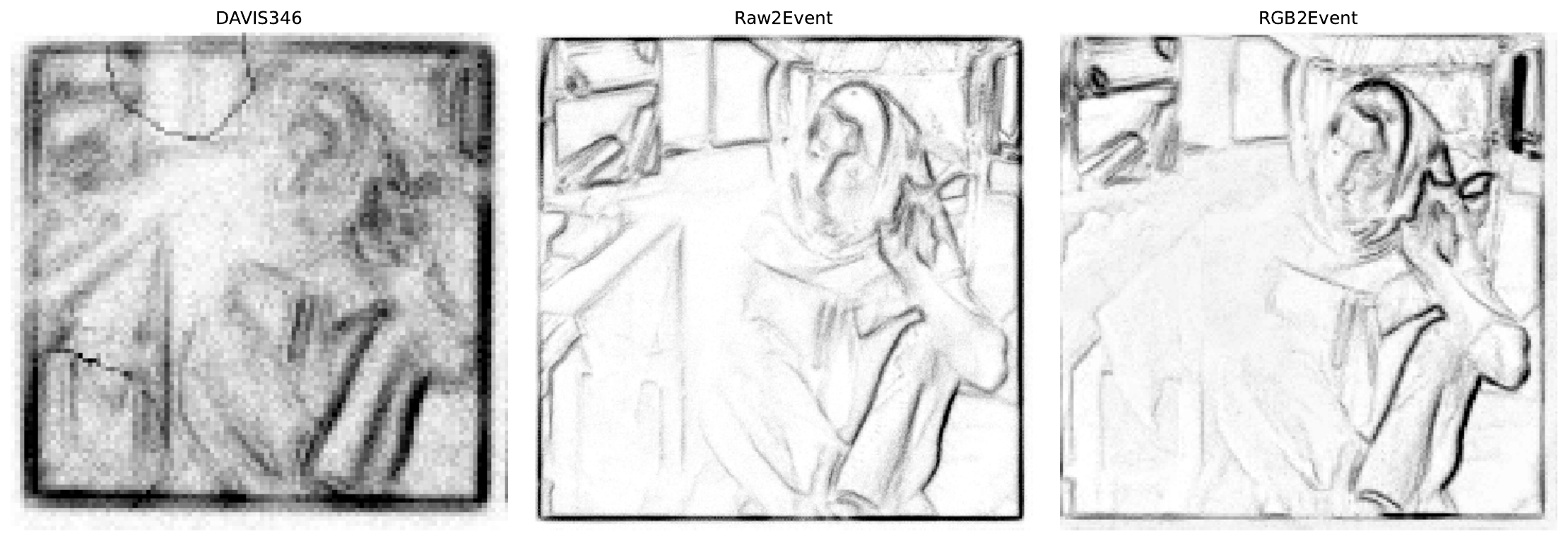}
    \caption{Heatmap (temporal accumulation) of events from DAVIS346, Raw2Event and RGB2Event, under HDR scene.}
    \label{fig:hdr_time_stack_heatmap}
\end{figure}

As illustrated in Fig.~\ref{fig:hdr_crop_frame}, we collected three types of event data - real events, simulated events generated from raw frames, and simulated events from RGB frames - all recording a static Barbara image displayed on the screen, while the cameras follow a circular trajectory. We used a DC light source to illuminate part of the image, creating a high dynamic range scene. Based on the temporal accumulation of events, it can be observed in Fig.~\ref{fig:hdr_time_stack_heatmap} that in bright areas, such as the edge of Barbara's right arm, the simulated events generated from raw frames preserve more spatial details compared to those generated from RGB frames. This advantage is further attributed to the higher pixel resolution and greater bit depth of raw data (10 bits vs. 8 bits) under identical acquisition settings.

\subsubsection{Autofocus Function}
\begin{figure}[htbp]
    \centering
    \includegraphics[width=0.9\linewidth, trim=0 0 0 400, clip]{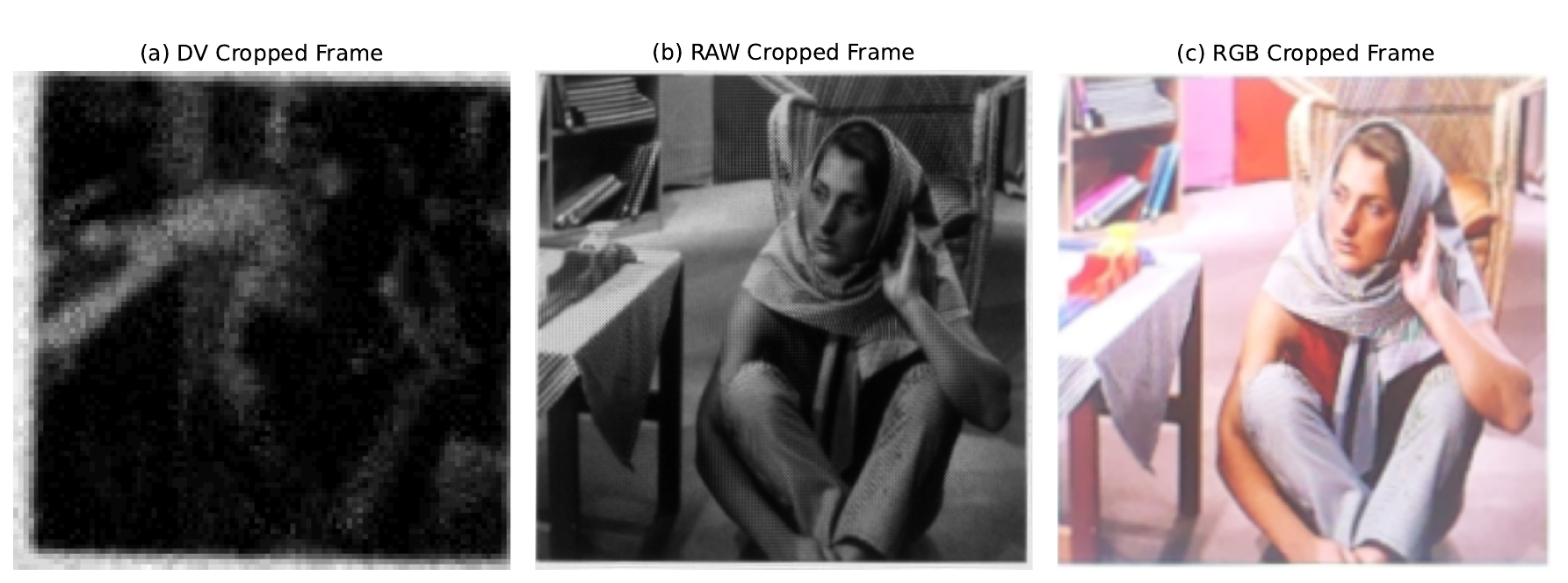}
    \caption{Views from frame sensor of DAVIS346 (defocused), raw and RGB sensor of Pi Cam.}
    \label{fig:focus_crop_frame}
\end{figure}

\begin{figure}[htbp]
    \centering
    \includegraphics[width=0.9\linewidth]{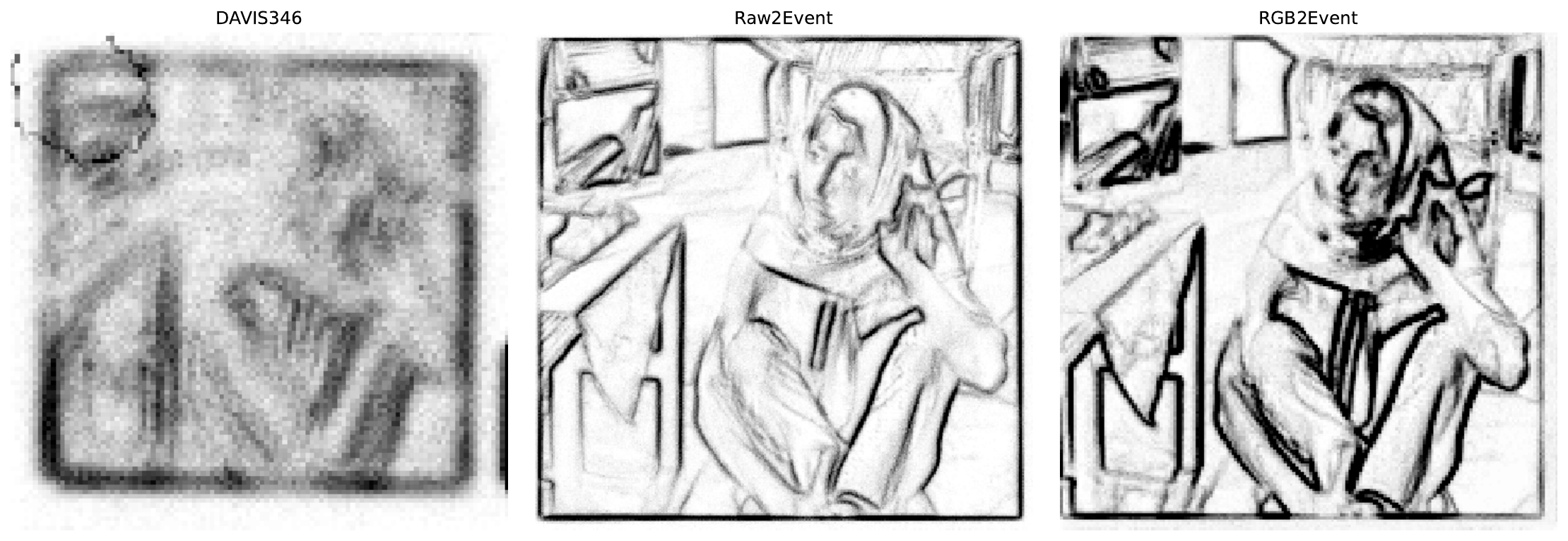}
    \caption{Heatmap of events from DAVIS346, Raw2Event and RGB2Event, showing the influence of focus.}
    \label{fig:focus_time_stack_heatmap}
\end{figure}

In a second qualitative study, we intentionally altered the distance between the image and the camera to test the robustness of different systems to focusing failures, as shown in Fig.~\ref{fig:focus_crop_frame}. In this scenario, the DAVIS sensor lost focus, while the Raspberry Pi camera maintained focus using its built-in autofocus feature. As shown in Fig.~\ref{fig:focus_time_stack_heatmap}, the DAVIS346 outputs low-quality events due to blurred gradients. Meanwhile, events generated from raw and RGB stream continue to have sharp event boundaries, highlighting the practical advantages of frame sensors with autofocus capabilities in low-cost environments.

\subsection{Event Behavior Across Tunable Parameters}
\label{sec:Event Behavior Across Tunable Parameters}

One of the core features of the Raw2Event system is its configurable event generation behavior. As introduced in Section~\ref{sec:User-Intuitive Parameters for Event Behavior Control}, users can adjust a set of parameters $k$ that govern event triggering rate, noise sensitivity, and timing stochasticity, etc. To systematically explore the effects of these parameters, we conducted a series of controlled experiments. Taking the calibrated $k$ values from Section~\ref{sec:Cross-Sensor Parameter Calibration} as the baseline, we varied one parameter at a time while keeping the others fixed, and converted the same raw image sequence into event streams.

\begin{figure*}[htbp]
    \centering
    \includegraphics[width=0.9\linewidth]{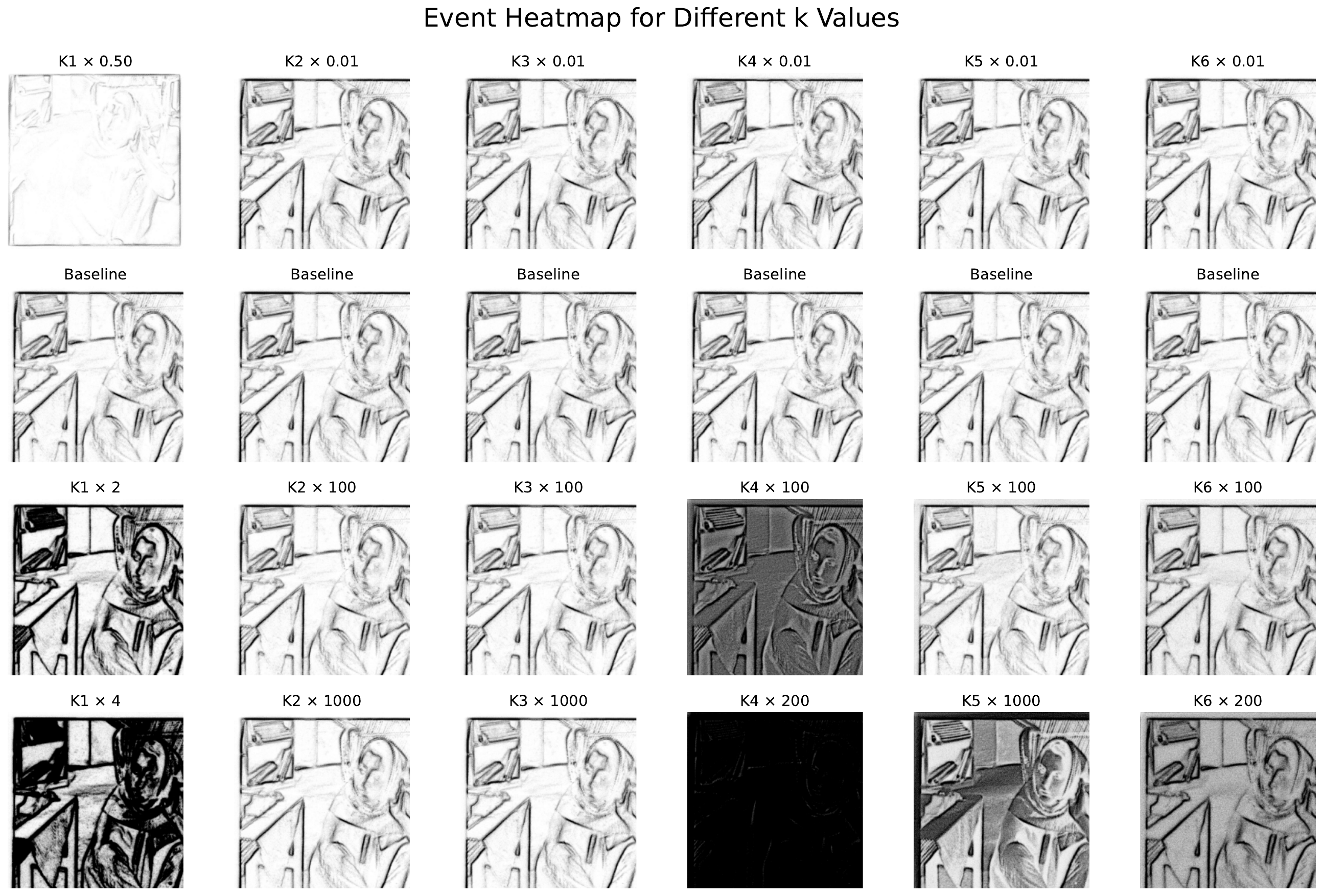}
    \caption{Heatmap of generated events under different $k$. Darker means more events.}
    \label{fig:heatmap_grid_plot_inv}
\end{figure*}

Figure~\ref{fig:heatmap_grid_plot_inv} shows spatial heatmaps of simulated events under different values of the contrast sensitivity parameter $k$. We observe that $k_1$ exhibits particularly high sensitivity: even small changes in its value lead to significant differences in event density. For high $k_1$ values (bottom two rows), the system generates a large number of events, including those in low-texture regions. In contrast, for low $k_1$ values (top row), events tend to cluster around high-contrast edges, and weak edges become less visible. This indicates that by adjusting a single parameter, Raw2Event can switch between ``high-sensitivity, high-noise" and ``low-sensitivity, edge-selective" modes of operation. Such flexibility allows researchers to synthesize event streams tailored to their specific application needs.

It is worth noting that the noise-related parameters $k_4$ and $k_6$ are also sensitive. Unless one aims to simulate highly noisy conditions, these values should be kept low. In fact, setting these parameters to zero can approximate an ideal, near-noiseless event sensor. Although increasing noise parameters may enhance edge visibility, this effect is not due to better detection, but rather to the implicit lowering of the event threshold, causing events in nearly static regions. Such behavior, while visually sharp, does not reflect the working principles of event cameras.

\begin{figure*}[htbp]
    \centering
    \includegraphics[width=0.9\linewidth]{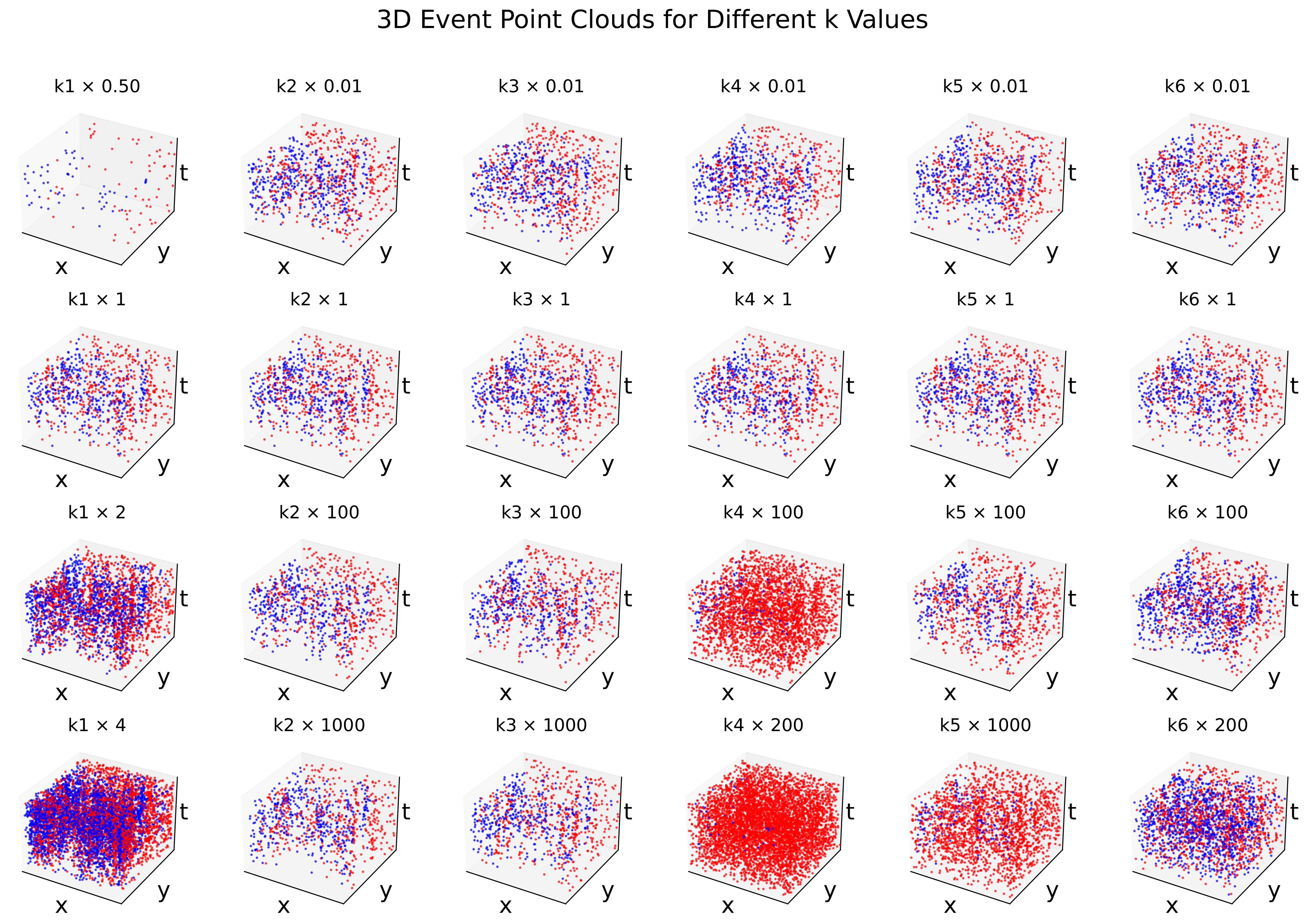}
    \caption{3D point cloud visualizations ($x$,$y$,$t$) of generated events under different $k$. Red points show ON events and blue points show OFF events.}
    \label{fig:event_point_clouds}
\end{figure*}

Figure~\ref{fig:event_point_clouds} presents the spatiotemporal distribution of generated events as 3D point clouds in the $(x,y,t)$ domain. As expected, increasing the values of parameters modeling imperfections in hardware, such as $k_4$ (noise) and $k_5$ (leakage), results in significantly more background events, particularly an overproduction of positive polarity events. These events do not correspond to meaningful scene changes but instead arise from simulated hardware-level noise. 

\subsection{Real-Time Performance on Embedded Platforms}
Despite the optimization using NumPy, the computational power of embedded devices remains the primary bottleneck for event generation. At a resolution of 346×260 (comparable to DAVIS346), our implementation on Raspberry Pi 4 achieves 10 FPS, while Raspberry Pi 5 achieves 33 FPS.

However, in scenarios where real-time display is not required, the system supports saving frames asynchronously, followed by offline event generation using PyTorch and GPU, which is useful for dataset collection and event generation performance benchmarking. In this case, we reached up to 80 FPS for the experiments in Section~\ref{sec:Event Behavior Across Tunable Parameters}, using a desktop with NVIDIA RTX 4070 Ti Super.

\section{Conclusion and Future Work}
\label{ch:conclusionandfuturework}

In this work, we presented Raw2Event, a cost-efficient, configurable system for simulating event-based vision using raw frame cameras. By leveraging raw sensor access, real-time event conversion, and flexible parameter modeling, Raw2Event offers a cheap alternative to physical event cameras for prototyping and low-cost deployment. Through comprehensive experiments, we also show that Raw2Event enables fine-grained control over noise, event responsiveness and randomness. Our design demonstrates that under reasonable lighting and motion conditions, it is possible to generate realistic event data streams from frame sensors, with advantages of autofocus and higher dynamic range than an RGB-based event generator.

Building on the demonstrated utility, we identify several promising avenues for extending and refining the concept of Raw2Event introduced in this paper.

The current Raw2Event framework takes grayscale images as input. However, with access to raw Bayer-pattern color information, it is feasible to extend the system to support color event streams, allowing for multi-channel event learning and richer scene understanding. This could further model color event sensors. 

In addition to supporting real-time operation on Raspberry Pi, the Raw2Event system can be further deployed on edge AI hardware (e.g., NVIDIA Jetson, AMD FPGAs), enabling even lower latency. This would allow Raw2Event to serve as a deployable solution for battery-powered, real-time edge applications where event-based vision is needed.

Another promising direction is self-adaptable event generation. Future extensions of Raw2Event could leverage the built-in feedback mechanisms of ISP to dynamically adjust exposure parameters in response to scene conditions. The conversion pipeline should be able to distinguish the light change caused by scene or by the control loop. This would allow the system to adaptively regulate sensitivity and extend the effective dynamic range in real time. Furthermore, by introducing learnable parameters that respond to the characteristics of the generated events (e.g. event density, polarity balance, and noise levels), the system could automatically adapt its behavior to suit varying lighting and motion environments.

By offering a cheap solution to transfer from frame-based to event-based sensing, Raw2Event opens up new possibilities for researchers, developers, and educators to explore event-driven perception without the need for expensive hardware. We hope it can serve as a useful tool to advance neuromorphic vision research and practical adoption.

\section*{Acknowledgments}
This work is partially supported by the Flemish Government via the Flanders AI Research Initiative (FAIR).

\bibliography{ref}

\begin{thebibliography}{10}
\providecommand{\url}[1]{#1}
\csname url@samestyle\endcsname
\providecommand{\newblock}{\relax}
\providecommand{\bibinfo}[2]{#2}
\providecommand{\BIBentrySTDinterwordspacing}{\spaceskip=0pt\relax}
\providecommand{\BIBentryALTinterwordstretchfactor}{4}
\providecommand{\BIBentryALTinterwordspacing}{\spaceskip=\fontdimen2\font plus
\BIBentryALTinterwordstretchfactor\fontdimen3\font minus \fontdimen4\font\relax}
\providecommand{\BIBforeignlanguage}[2]{{%
\expandafter\ifx\csname l@#1\endcsname\relax
\typeout{** WARNING: IEEEtran.bst: No hyphenation pattern has been}%
\typeout{** loaded for the language `#1'. Using the pattern for}%
\typeout{** the default language instead.}%
\else
\language=\csname l@#1\endcsname
\fi
#2}}
\providecommand{\BIBdecl}{\relax}
\BIBdecl

\bibitem{gallegoEventbasedVisionSurvey2022}
G.~Gallego, T.~Delbruck, G.~Orchard, C.~Bartolozzi, B.~Taba, A.~Censi, S.~Leutenegger, A.~Davison, J.~Conradt, K.~Daniilidis, and D.~Scaramuzza, ``Event-based {{Vision}}: {{A Survey}},'' \emph{IEEE Transactions on Pattern Analysis and Machine Intelligence}, vol.~44, no.~1, pp. 154--180, Jan. 2022.

\bibitem{huV2eVideoFrames2021}
Y.~Hu, S.-C. Liu, and T.~Delbruck, ``V2e: {{From Video Frames}} to {{Realistic DVS Events}},'' Apr. 2021.

\bibitem{zhangV2CEVideoContinuous2024}
Z.~Zhang, S.~Cui, K.~Chai, H.~Yu, S.~Dasgupta, U.~Mahbub, and T.~Rahman, ``{{V2CE}}: {{Video}} to {{Continuous Events Simulator}},'' Apr. 2024.

\bibitem{linDVSVoltmeterStochasticProcessBased2022}
S.~Lin, Y.~Ma, Z.~Guo, and B.~Wen, ``{{DVS-Voltmeter}}: {{Stochastic Process-Based Event Simulator}} for~{{Dynamic Vision Sensors}},'' in \emph{Computer {{Vision}} -- {{ECCV}} 2022}, S.~Avidan, G.~Brostow, M.~Ciss{\'e}, G.~M. Farinella, and T.~Hassner, Eds.\hskip 1em plus 0.5em minus 0.4em\relax Cham: Springer Nature Switzerland, 2022, pp. 578--593.

\bibitem{jamilReviewImageQuality2024}
S.~Jamil, ``Review of {{Image Quality Assessment Methods}} for {{Compressed Images}},'' \emph{Journal of Imaging}, vol.~10, no.~5, p. 113, May 2024.

\bibitem{Deblurring}
X.~Zhang and L.~Yu, ``Unifying motion deblurring and frame interpolation with events,'' in \emph{2022 IEEE/CVF Conference on Computer Vision and Pattern Recognition (CVPR)}, 2022, pp. 17\,744--17\,753.

\bibitem{hdr}
\BIBentryALTinterwordspacing
N.~Messikommer, S.~Georgoulis, D.~Gehrig, S.~Tulyakov, J.~Erbach, A.~Bochicchio, Y.~Li, and D.~Scaramuzza, ``{ Multi-Bracket High Dynamic Range Imaging with Event Cameras },'' in \emph{2022 IEEE/CVF Conference on Computer Vision and Pattern Recognition Workshops (CVPRW)}.\hskip 1em plus 0.5em minus 0.4em\relax Los Alamitos, CA, USA: IEEE Computer Society, Jun. 2022, pp. 546--556. [Online]. Available: \url{https://doi.ieeecomputersociety.org/10.1109/CVPRW56347.2022.00070}
\BIBentrySTDinterwordspacing

\bibitem{lichtsteiner128times1281202008}
P.~Lichtsteiner, C.~Posch, and T.~Delbruck, ``A 128{\textbackslash}times 128 120 {{dB}} 15 {$M$}s {{Latency Asynchronous Temporal Contrast Vision Sensor}},'' \emph{IEEE Journal of Solid-State Circuits}, vol.~43, no.~2, pp. 566--576, Feb. 2008.

\bibitem{Gehrig2024Nature}
\BIBentryALTinterwordspacing
D.~Gehrig, A.~Loquercio, M.~Gehrig, and D.~Scaramuzza, ``Low-latency automotive vision with high-speed, low-power event cameras,'' \emph{Nature}, vol. 628, no. 8003, pp. 522--528, 2024. [Online]. Available: \url{https://www.nature.com/articles/s41586-024-07409-w}
\BIBentrySTDinterwordspacing

\bibitem{mentastiEventBasedObjectDetection2022}
S.~Mentasti, A.~W. Kambale, and M.~Matteucci, ``Event-{{Based Object Detection}} and {{Tracking}} - {{A Traffic Monitoring Use Case}},'' in \emph{Image {{Analysis}} and {{Processing}}. {{ICIAP}} 2022 {{Workshops}}}, ser. Lecture {{Notes}} in {{Computer Science}}, P.~L. Mazzeo, E.~Frontoni, S.~Sclaroff, and C.~Distante, Eds.\hskip 1em plus 0.5em minus 0.4em\relax Cham: Springer International Publishing, 2022, pp. 95--106.

\bibitem{falangaDynamicObstacleAvoidance2020}
D.~Falanga, K.~Kleber, and D.~Scaramuzza, ``Dynamic obstacle avoidance for quadrotors with event cameras,'' \emph{Science Robotics}, vol.~5, no.~40, p. eaaz9712, Mar. 2020.

\bibitem{jiangMixedFrameEventDriven2019}
Z.~Jiang, P.~Xia, K.~Huang, W.~Stechele, G.~Chen, Z.~Bing, and A.~Knoll, ``Mixed {{Frame-}}/{{Event-Driven Fast Pedestrian Detection}},'' in \emph{2019 {{International Conference}} on {{Robotics}} and {{Automation}} ({{ICRA}})}, May 2019, pp. 8332--8338.

\bibitem{imagenet}
J.~Deng, W.~Dong, R.~Socher, L.-J. Li, K.~Li, and L.~Fei-Fei, ``Imagenet: A large-scale hierarchical image database,'' in \emph{2009 IEEE Conference on Computer Vision and Pattern Recognition}, 2009, pp. 248--255.

\bibitem{muegglerEventCameraDatasetSimulator2017}
E.~Mueggler, H.~Rebecq, G.~Gallego, T.~Delbruck, and D.~Scaramuzza, ``The {{Event-Camera Dataset}} and {{Simulator}}: {{Event-based Data}} for {{Pose Estimation}}, {{Visual Odometry}}, and {{SLAM}},'' \emph{The International Journal of Robotics Research}, vol.~36, no.~2, pp. 142--149, Feb. 2017.

\bibitem{rebecqESIMOpenEvent2018}
H.~Rebecq, D.~Gehrig, and D.~Scaramuzza, ``{{ESIM}}: An {{Open Event Camera Simulator}},'' in \emph{Proceedings of {{The}} 2nd {{Conference}} on {{Robot Learning}}}.\hskip 1em plus 0.5em minus 0.4em\relax PMLR, Oct. 2018, pp. 969--982.

\bibitem{gehrigVideoEventsRecycling2020}
D.~Gehrig, M.~Gehrig, J.~{Hidalgo-Carri{\'o}}, and D.~Scaramuzza, ``Video to {{Events}}: {{Recycling Video Datasets}} for {{Event Cameras}},'' Apr. 2020.

\bibitem{zhuEventGANLeveragingLarge2019}
A.~Z. Zhu, Z.~Wang, K.~Khant, and K.~Daniilidis, ``{{EventGAN}}: {{Leveraging Large Scale Image Datasets}} for {{Event Cameras}},'' Dec. 2019.

\bibitem{nehviEventHandsRealTimeNeural2021}
J.~Nehvi, V.~Golyanik, F.~Mueller, H.-P. Seidel, M.~Elgharib, and C.~Theobalt, ``{{EventHands}}: {{Real-Time Neural 3D Hand Reconstruction}} from an {{Event Stream}},'' in \emph{{{CVPR Workshops}}}, 2021.

\bibitem{planamenteDA4EventBridgingSimtoReal2021}
M.~Planamente, C.~Plizzari, M.~Cannici, M.~Ciccone, F.~Strada, A.~Bottino, M.~Matteucci, and B.~Caputo, ``{{DA4Event}}: Towards bridging the {{Sim-to-Real Gap}} for {{Event Cameras}} using {{Domain Adaptation}},'' Oct. 2021.

\bibitem{guReliableEventGeneration2024}
D.~Gu, J.~Li, L.~Zhu, Y.~Zhang, and J.~S. Ren, ``Reliable {{Event Generation With Invertible Conditional Normalizing Flow}},'' \emph{IEEE Transactions on Pattern Analysis and Machine Intelligence}, vol.~46, no.~2, pp. 927--943, Feb. 2024.

\bibitem{serrano-gotarredonaPokerDVSMNISTDVSTheir2015}
T.~{Serrano-Gotarredona} and B.~{Linares-Barranco}, ``Poker-{{DVS}} and {{MNIST-DVS}}. {{Their History}}, {{How They Were Made}}, and {{Other Details}},'' \emph{Frontiers in Neuroscience}, vol.~9, Dec. 2015.

\bibitem{orchardConvertingStaticImage2015}
G.~Orchard, A.~Jayawant, G.~K. Cohen, and N.~Thakor, ``Converting {{Static Image Datasets}} to {{Spiking Neuromorphic Datasets Using Saccades}},'' \emph{Frontiers in Neuroscience}, vol.~9, Nov. 2015.

\bibitem{liCIFAR10DVSEventStreamDataset2017}
H.~Li, H.~Liu, X.~Ji, G.~Li, and L.~Shi, ``{{CIFAR10-DVS}}: {{An Event-Stream Dataset}} for {{Object Classification}},'' \emph{Frontiers in Neuroscience}, vol.~11, 2017.

\bibitem{cifar10}
A.~Krizhevsky and G.~Hinton, ``Learning multiple layers of features from tiny images,'' University of Toronto, Tech. Rep., 2009, technical Report.

\bibitem{kimNImageNetRobustFineGrained2021}
J.~Kim, J.~Bae, G.~Park, D.~Zhang, and Y.~M. Kim, ``N-{{ImageNet}}: {{Towards Robust}}, {{Fine-Grained Object Recognition With Event Cameras}},'' in \emph{Proceedings of the {{IEEE}}/{{CVF International Conference}} on {{Computer Vision}}}, 2021, pp. 2146--2156.

\bibitem{perotLearningDetectObjects2020}
E.~Perot, P.~{de Tournemire}, D.~Nitti, J.~Masci, and A.~Sironi, ``Learning to {{Detect Objects}} with a 1 {{Megapixel Event Camera}},'' Dec. 2020.

\bibitem{plizzariE2GOMOTIONMotionAugmented2022}
C.~Plizzari, M.~Planamente, G.~Goletto, M.~Cannici, E.~Gusso, M.~Matteucci, and B.~Caputo, ``E2({{GO}}){{MOTION}}: {{Motion Augmented Event Stream}} for {{Egocentric Action Recognition}},'' in \emph{Proceedings of the {{IEEE}}/{{CVF Conference}} on {{Computer Vision}} and {{Pattern Recognition}}}, 2022, pp. 19\,935--19\,947.

\bibitem{chakravarthiRecentEventCamera2024}
B.~Chakravarthi, A.~A. Verma, K.~Daniilidis, C.~Fermuller, and Y.~Yang, ``Recent {{Event Camera Innovations}}: {{A Survey}},'' Aug. 2024.

\bibitem{picamera2manual}
{Raspberry Pi Ltd.}, \emph{Picamera2 Documentation}, Raspberry Pi Ltd., 2024, \url{https://datasheets.raspberrypi.com/camera/picamera2-manual.pdf}.

\bibitem{wangAprilTag2Efficient2016}
J.~Wang and E.~Olson, ``{{AprilTag}} 2: {{Efficient}} and robust fiducial detection,'' in \emph{2016 {{IEEE}}/{{RSJ International Conference}} on {{Intelligent Robots}} and {{Systems}} ({{IROS}})}, Oct. 2016, pp. 4193--4198.

\bibitem{optuna_2019}
T.~Akiba, S.~Sano, T.~Yanase, T.~Ohta, and M.~Koyama, ``Optuna: A next-generation hyperparameter optimization framework,'' in \emph{Proceedings of the 25th {ACM} {SIGKDD} International Conference on Knowledge Discovery and Data Mining}, 2019.

\end{thebibliography}


\end{document}